\documentclass[twoside,11pt]{article}

\usepackage[abbrvbib, preprint]{jmlr2e}

\usepackage{hyperref}
\usepackage{xcolor}
\usepackage{amssymb}
\usepackage{amscd}
\usepackage{amsmath}
\usepackage{tikz}
\usepackage{tikzit}
\tikzstyle{new style 0}=[fill={rgb,255: red,10; green,26; blue,255}, draw=black, shape=circle]
\tikzstyle{new style 1}=[fill=yellow, draw=black, shape=circle]

\tikzstyle{redline}=[-, draw=red]
\tikzstyle{blackline}=[-]
\tikzstyle{grayarrow}=[draw={rgb,255: red,141; green,141; blue,141}, ->]
\tikzstyle{blackarrow}=[draw=black, ->]

\usepackage{tikz-cd}
\usetikzlibrary{calc}

\tikzset{none/.style={thick}}
\tikzset{simple/.style={thick}}

\tikzset{
    ncbar angle/.initial=90,
    ncbar/.style={
        to path=(\tikztostart)
        -- ($(\tikztostart)!#1!\pgfkeysvalueof{/tikz/ncbar angle}:(\tikztotarget)$) 
        -- ($(\tikztotarget)!($(\tikztostart)!#1!\pgfkeysvalueof{/tikz/ncbar angle}:(\tikztotarget)$)!\pgfkeysvalueof{/tikz/ncbar angle}:(\tikztostart)$) 
           \tikztonodes
        -- (\tikztotarget) 
    },
    ncbar/.default=0.5cm,
}

\newtheorem{Theorem}{Theorem}[section]
\newtheorem{Conjecture}{Conjecture}[section]

\newcommand{\dataset}{{\cal D}}
\newcommand{\fracpartial}[2]{\frac{\partial #1}{\partial  #2}}
\newcommand{\Loss}{\mathcal{L}}

\newcommand{\maxx}{\mathbf{max}}
\newcommand{\minn}{\mathbf{min}}

\ShortHeadings{Neural Teleportation}{Armenta and Judge et al}
\firstpageno{1}

\begin{document}

\title{Neural Teleportation}

\author{\name Marco Armenta $^{1,2}$ \email marco.armenta@usherbrooke.ca \\
       \addr $^{1}$ Department of Computer Science\\
       Université de Sherbrooke\\
       Sherbrooke, QC J1K 2R1, Canada. \\
       $^{2}$ Department of Mathematics\\
       Université de Sherbrooke\\
       Sherbrooke, QC J1K 2R1, Canada.
       \AND
       \name Thierry Judge $^{1}$ \email thierry.judge@usherbrooke.ca \\
       \name Nathan Painchaud $^{1}$ \email nathan.painchaud@usherbrooke.ca \\
       \name Youssef Skandarani $^{3}$ \email youssef\_skandarani@etu.u-bourgogne.fr \\
       \addr $^{3}$ Université de Bourgogne Franche-Comte\\
       Dijon, France.
       \AND
       \name Carl Lemaire $^{1}$ \email carl.lemaire@usherbrooke.ca \\
       \name Gabriel Gibeau Sanchez $^{1}$ \email philippe.spino@usherbrooke.ca \\
       \name Philippe Spino $^{1}$ \email gabriel.gibeau.sanchez@usherbrooke.ca \\
       \name Pierre-Marc Jodoin $^{1}$ \email pierre-marc.jodoin@usherbrooke.ca
       }

\editor{-}

\maketitle

\begin{abstract}%   <- trailing '%' for backward compatibility of .sty file
%In this paper we explore the question '\textit{what happens when we apply an isomorphism of quiver representations to a neural network?}'. There is little information concerning this question, as only a very small type of isomorphisms have been explored.

In this paper, we explore a process called neural teleportation, a mathematical consequence of applying quiver representation theory to neural networks. Neural teleportation ``\textit{teleports}" a network to a new position in the weight space and preserves its function. This phenomenon comes directly from the definitions of representation theory applied to neural networks and it turns out to be a very simple operation that has remarkable properties.

%This concept generalizes the notion of positive scale invariance of ReLU networks to any network with any activation functions and any architecture.

We shed light on surprising and counter-intuitive consequences neural teleportation has on the loss landscape. In particular, we show that teleportation can be used to explore loss level curves, that it changes the local loss landscape, sharpens global minima and boosts back-propagated gradients at any moment during the learning process. Our results can be reproduced with the code available here: \href{https://github.com/vitalab/neuralteleportation}{\color{blue}{github.com/vitalab/neuralteleportation}}.

%From these observations, we demonstrate that teleportation accelerates training when used during initialization regardless of the model, its activation function, the loss function, and the training data. 

\end{abstract}

\begin{keywords}
  quiver representations, teleportation, isomorphisms, positive scale invariance.
\end{keywords}

\section{Introduction}

Despite years of research, our theoretical understanding of neural networks, their loss landscape and their behavior during training and testing is still limited. A recent novel theoretical analysis of neural networks using quiver representation theory given by \cite{ArmentaJodoin20} has been introduced, where the algebraic and combinatorial nature of neural networks is exposed. Among other things, the authors present a by-product of representing neural networks through the lens of quiver representation theory, {\em i.e.}, the notion of \textit{neural teleportation}.

This is the mathematical foundation that explains why practitioners of deep learning have observed the property of scale invariance for the very particular case of neural networks with positive scale invariant activation functions, see for example the work of~\cite{Neyshabur15}. Nevertheless, this type of invariance has not been studied or observed in its full generality, for example, across batch norm layers, residual connections, concatenation layers or activation functions other than ReLU. Even more, it has been claimed, for example, by \cite{Meng18}, that there is no scale invariance across residual connections. This is because it is not until a neural network is drawn as a quiver that the invariance of the network function under isomorphisms across any architecture becomes obvious (see appendix for an illustration of this). Positive scale invariance also restricts the type of scaling factor reducing them to positive real numbers and positive scale invariant activation functions, while neural teleportation allows to teleport any neural architecture with any non-zero scaling factor.

In this work, %we do not intent to beat the state-of-the-art performance of any training method for deep neural networks. We intent only 
we intend to exhibit empirical evidence that using quiver representation theoretic concepts produces measurable changes on the behaviours of deep neural networks. 
Due to its mathematical nature, neural teleportation is an intrinsic property of every neural network, and so it is independent of the architecture, the activation functions and even the task at hand or the data. Here, we perform extensive experiments on classification tasks with feedforward neural networks with different activation functions and scaling factor sampling.

%It is thanks to the nature of the quiver approach that the statements and proofs in this paper may look elementary. Once everything is in the language of quiver representations the proofs are straightforward.

%The application of isomorphisms of quiver representations to neural networks allows for a vast liberty on the degrees of freedom to teleport a neural network. Unlike positive scale invariance that puts restrictions on both the architecture (only positive scale invariant activation functions, not across residual or concatenation or batchnorm layers) and the scaling parameters. 

%We present here extensive examples where neural teleportation exhibits unexpected results.

%a theoretical analysis of the mathematical approach of quiver representations. 

%We stress out that the process of teleporting a neural network works for any type of neural network independently of the architecture, activation functions and even the task. 

%We also mention that teleportation is a very general intrinsic property of every neural network in which even complex numbers can be used at any stage without changing the output of the neural network after teleportation.

%Neural networks have achieved very high performance across many areas of machine learning \cite{Goodfellow-et-al-2016,LeCun15,raghu2020survey}. 
 % (see \cite{ArmentaJodoin20} page 19).  

As will be explained later, neural teleportation is the mathematical consequence of applying the concept of {\em isomorphisms of quiver representations} to neural networks.  This process has the unique property of changing the weights and the activation functions of a network while, at the same time, preserving its function, {\em i.e.}, a teleported network makes the same predictions for the same input values as the original, non-teleported network. %(i.e. the network before and after changing of weight and activation function .   

Isomorphisms of quiver representations have already been used on neural networks, often unbeknownst to the authors, through the concept of {\em positive scale invariance} (also called {\em positive homogeneity}) of some activation functions, see~\cite{Badrinarayanan15, Dinh17, Meng18, Neyshabur15}. However, representation theory lays down the mathematical foundations of this phenomenon and explains why this has been observed only on networks with positive scale invariant activation functions. Following the quiver representation theoretic approach to neural networks it becomes clear that any neural network can be teleported with an isomorphism, as opposed to what is remarked in the literature. Namely, it has been claimed that there is no scale invariance across residual connections \citep{Meng18} or on the parameters $\beta$ and $\gamma$ for batch norm layers \citep{Arora19}. We explain (see appendix for illustrations on residual connections and batchnorm layers) how the quiver approach is essential to apply teleportation to any architecture.

The concept of positive scale invariance derives from the fact that one can choose a \textit{positive} number $c$ for each hidden neuron of a \textit{ReLU} network, multiply every incoming weight to that neuron by $c$ and divide every outgoing weight by $c$ and still have the same network function. Note that models like maxout networks \citep{Goodfellow13}, leaky rectifiers \citep{He15} and deep linear networks \citep{Saxe13} are also positive scale invariant.
This concept in previous works is always restricted to positive scale invariant activation functions and to positive scaling factors. However, the generality in which quiver representation theory describes neural networks allows to teleport any architecture, with any activation function and any non-zero scaling factor.  %, see \cite{ArmentaJodoin20} Corollary 4.10, page 14. 

\begin{figure}
\includegraphics[width=.98\linewidth]{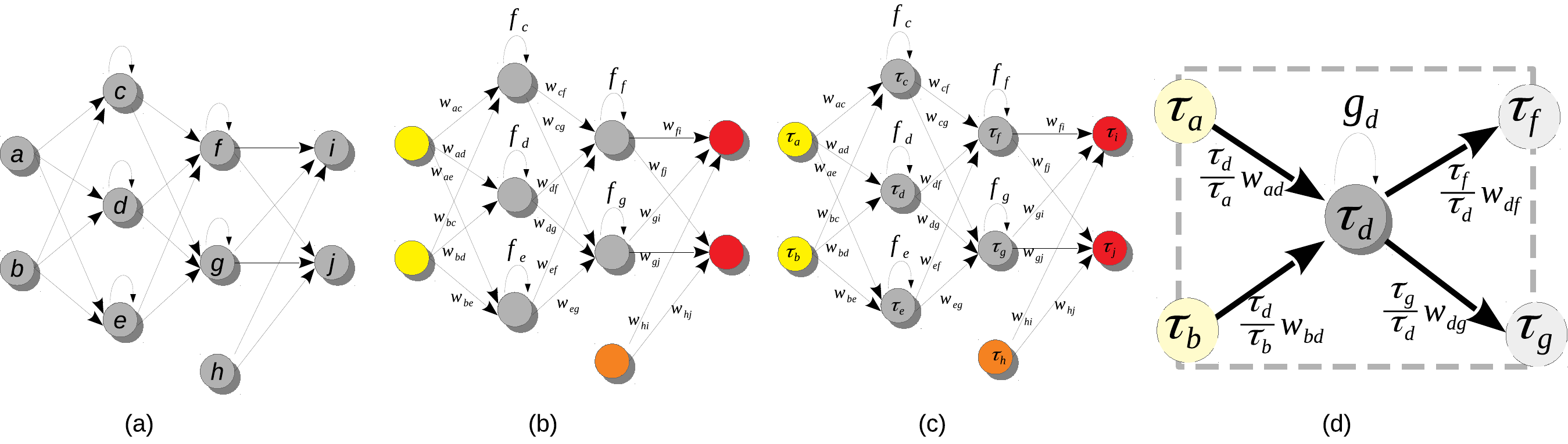}
\caption{(a) Network quiver $Q$ as introduced by~\cite{ArmentaJodoin20}. (b) Neural network based on $Q$ with weights $W$ and activation functions $f$. Input, hidden, bias, and output neurons are in yellow, gray, orange and red, respectively. (c) Same neural network but with a change of basis (CoB) $\tau_\epsilon$ at each neuron $\epsilon$. (d) Neural teleportation of the weights attached to neuron $d$. The resulting activation function is $g_d(x)=\tau_d f_d(x/\tau_d)$. 
} \label{fig:Net_Quiv_Ex}
\end{figure}

%Positive scale invariance is often seen as some sort of trick to change the weights of a ReLU network without affecting the network function.  

In this paper, we show that neural teleportation is more than a trick, as it has concrete consequences on the loss landscape and the network optimization.  We account for various theoretical and empirical consequences that teleportation has on neural networks.%, but also has an impact on the shape of the loss landscape. 

%To picture what neural teleportation is, lets consider $D$ a set of labeled data, $W$ the weights of a neural network and $f$ its activation function (ReLU, tanh, etc).  As shown in Fig.~\ref{fig:lossLandscape}, $W$ can be seen as a point lying on the slope of a loss landscape ${\cal L}_{f,W}(D)$ that a gradient descent ought to iteratively climb down.  % : $W^{[i+1]}\leftarrow W^{[i]} - \eta \nabla{{\cal L}_{f,W^{[i]}}(D)}$.  
%In general, changing the content of $W$ and/or the function $f$ (say, from ReLU to tanh) while keeping $D$ fixed has the effect of changing the loss value ${\cal L}_{f,W}(D)$.  %After all, it is what gradient descent is all about: changing $W$ in order to reduce the loss.

%Neural teleportation is a process that transforms (or "teleport") the weights $W$ into a new set of weights $V$ through the application of a to-be-defined {\em change of basis (CoB)}.  If the activation function stays the same, $V$ can be seen as a different point in the same loss landscape $g$ (c.f. figure\ref{fig:lossLandscape}[left]).  However, depending on the CoB, neural teleportation can change the activation functions and thus create a new loss landscape (c.f. figure\ref{fig:lossLandscape}[right]).

%By its very nature, neural teleportation preserves the network function : i.e. ${\cal L}_{f,W}(D)\equiv {\cal L}_{g,V}(D)$.   In this paper, 

Our findings can be summarized as follows:
\begin{enumerate}
    \item Neural teleportation can be used to explore loss level curves;
    \item {\em Micro-teleportation vectors} have the property of being rigorously perpendicular to back-propagated gradients computed with any kind and amount of labeled data, even random data with random labels; 
    %\item[3.] We show that simply changing the weights of a network by a factor equal to the distance between the network and a teleportation is not enough to outperform a training without teleportation.
    \item Neural teleportation changes the flatness of the loss landscape; %We show how the change of basis affects the flatness of the loss landscape of teleportations by using two types of samplings for the change of basis;\vspace{-0.2cm}
    \item The back-propagated gradients of a teleported network scale with respect to the scaling factor;%, see Theorem \ref{thm:grad_tel}.\vspace{-0.2cm}
     %As a consequence of the previous points, 
    \item Randomly teleporting a network before training speeds up gradient descent (with and without momentum). Actually, we also found that \textit{one} teleportation can accelerate training even when used at the middle of the training. % and this depends on the CoB-range (to be introduced in the next section).
    %\item Teleportation reveals the presence of periodic patterns in the loss landscape of neural networks;
\end{enumerate}

%-------------------------------------------------------------------------

%\begin{figure*}
%\begin{center}
%\end{center}
%   \caption{The angle between the line between $W$ and $TW$ and the gradient. The change of basis range is (from left to right) 0.1, 0.01 and 0.001, and we computed this for a ResNet on the cifar100 data set.}
%\label{fig:micro2}
%\end{figure*}

\section{Neural teleportation}
\label{sec:NT}

In this section, we explain what neural teleportation is and the implications it has on the loss landscape and the back-propagated gradients. For more details on the theoretical interpretation of neural networks according to quiver representation theory, please refer to the work of \cite{ArmentaJodoin20}. %Here we will be working over the field of real numbers $\R$.

\subsection{Isomorphisms and Change of Basis (CoB)}

Neural networks are often pictured as oriented graphs.  \cite{ArmentaJodoin20} show that neural network graphs are a specific kind of quiver with a loop at every hidden node. They call these graphs \textit{network quivers} (c.f. Fig.~\ref{fig:Net_Quiv_Ex}(a)).  They also mention that neural networks, as they are generally defined, are network quivers with a weight assigned to every edge and an activation function at every loop (c.f. Fig.~\ref{fig:Net_Quiv_Ex}(b)).

According to representation theory, two quiver representations are equivalent if an {\em isomorphism} exists between the two. Mathematically, neural networks are quiver representations with activation functions, and this allows to apply isomorphisms of quiver representations directly to every neural network independently of the architecture or the task at hand.

%Since neural networks are a type of quiver representation, isomorphisms of quiver representations also applies to them. 

Isomorphisms are given by sets of non-zero real numbers subject to some conditions. One such set of non-zero numbers is called a \textit{change of basis} (CoB), see for example~\cite{Assem06}, where each node of the quiver is assigned a non-zero number.  In order to apply an isomorphism to a neural network, each neuron $\epsilon$ must be assigned a CoB represented by $\tau_\epsilon \in I\!\!R^{\mbox{\tiny $\not =$0}}$ in Fig.~\ref{fig:Net_Quiv_Ex}(c).  

However, isomorphisms of quiver representations are very general and they may break implementations of neural network layers.  For example, applying an isomorphism to a convolutional layer may not result in a convolutional layer. %often result into a fully-connected layer instead of a convolution. 
But, in any case, we will always obtain a network with the exact same network function. It was proved by \cite{ArmentaJodoin20} that these isomorphisms can be restricted to preserve the implementation of neural network layers for any architecture. Neural teleportation is the process of applying one of such particular isomorphisms of quiver representations to neural networks.

The conditions over the CoB to produce a teleportation of a neural network are the following~\citep{ArmentaJodoin20} :

%The process of applying an isomorphism % (subject to some conditions),
%to a neural net is called {\em neural teleportation}.  For a teleportation to be valid, the CoB must comply to the following four conditions: 
\begin{enumerate}
    \item The CoB of every input, output and bias neuron must be equal to $1$ ({\em i.e.} $\tau_a,\tau_b,\tau_h,\tau_i,\tau_j=1$ in Fig.~\ref{fig:Net_Quiv_Ex}(c)).
    \item Neurons $k,l$ connected by a residual connection should have the same CoB : $\tau_k=\tau_l$. See appendix for details.
    \item For convolutional layers, the neurons of a given feature map should share the same CoB. 
    \item For batch norm layers, a CoB must be assigned to parameters $\beta$ and $\gamma$, but not to the mean and variance. See appendix for details.
    \item The CoBs of neurons connected by a dense connection are obtained by concatenating those in its input layers.
\end{enumerate}

Condition 1 is the \textit{isomorphism condition}. Any two networks related by a CoB satisfying this condition are said to be \textit{isomorphic}.  Applying the notion of isomorphism of quiver representation to neural networks leads to the following theorem:% This comes from the following result by \cite{ArmentaJodoin20}. %Whenever two neural networks are related by an isomorphism (i.e., by a CoB satisfying condition 1), they have the same network function, as proved in \cite{ArmentaJodoin20}.\vspace{-0.1cm}

\begin{Theorem} \citep{ArmentaJodoin20}  \label{thm:isom-nets}
    Isomorphic neural networks have the same network function. 
\end{Theorem}

Said otherwise, despite having different weights and different activation functions, isomorphic neural networks return rigorously the {\em same predictions for the same inputs}.  It also means that they have {\em exactly} the same loss values (c.f. \cite{ArmentaJodoin20} for the proof).

Conditions 2 to 5 have to do with the architecture of the network. They ensure that the produced isomorphic networks share the same architecture (again c.f. \cite{ArmentaJodoin20} for more details).  These conditions ensure that the teleportation of a residual connection remains a residual connection (condition 2), the teleportation of a conv layer remains a conv layer (condition 3), the teleportation of a batch norm remains a batch norm (condition 4) and the teleportation of a dense layer remains a dense layer (condition 5). Please note that ~\cite{Meng18} introduced a concept similar to condition 3.

%Neural teleportation, given by a CoB satisfying the four conditions, produces a neural network with the same network function and the same programmed architecture.

\subsection{Teleporting a neural network}

Neural teleportation is a process by which the weights $W$ and the activation functions $f$ of a neural network are converted to a new set of weights $V$, and a new set of activation functions $g$.  From a practical standpoint, this process is illustrated in Fig.~\ref{fig:Net_Quiv_Ex}(d).  

Considering $w_{ab}\in I\!\!R$ the weight of the connection from neuron $a$ to neuron $b$, and $\tau_a  \in I\!\!R^{\mbox{\tiny $\not =$0}}$ (resp. $\tau_b  \in I\!\!R^{\mbox{\tiny $\not =$0}}$) the CoB of neuron $a$ (resp. $b$).  The teleportation of that weight is simply:
\begin{eqnarray}
\label{eq:neuralTeleport}
v_{ab}=\frac{\tau_b}{\tau_a}w_{ab.}
\end{eqnarray}
To teleport an entire network, this operation is carried out for every weight of the network.  Note that positive homogeneity, see \cite{Badrinarayanan15, Dinh17, Meng18, Neyshabur15}, implies a similar operation but with the restriction of positive scaling factors.  In the case of batch norm layers, the parameter $\gamma$ is treated like a weight between two hidden neurons and $\beta$ as a weight starting from a bias neuron~\citep{ArmentaJodoin20}. As such, $\gamma$ and $\beta$ are teleported like any other weight using Eq.(\ref{eq:neuralTeleport}).

Neural teleportation also applies to activation functions. If $f_d$ is the activation function of neuron $d$ in Fig.~\ref{fig:Net_Quiv_Ex}(d), then the teleported activation is 
\begin{eqnarray} \label{eq:tel-act}
    g_d(x) = \tau_d \cdot  f_d\left(\frac{x}{\tau_d}\right).
\end{eqnarray}
This is a critical operation to make sure the pre- and post-teleported networks have the same function.  We can see that if  $\tau_d > 0$ and $f_d$ is positive scale invariant (like ReLU)  then $g_d(x) = \tau_d f_d(x / \tau_d) = \tau_d / \tau_d f_d(x) = f_d(x)$, and so positive scale invariance is a consequence of neural teleportation.

\begin{figure}[t]
\begin{center}
\includegraphics[width=.75\linewidth]{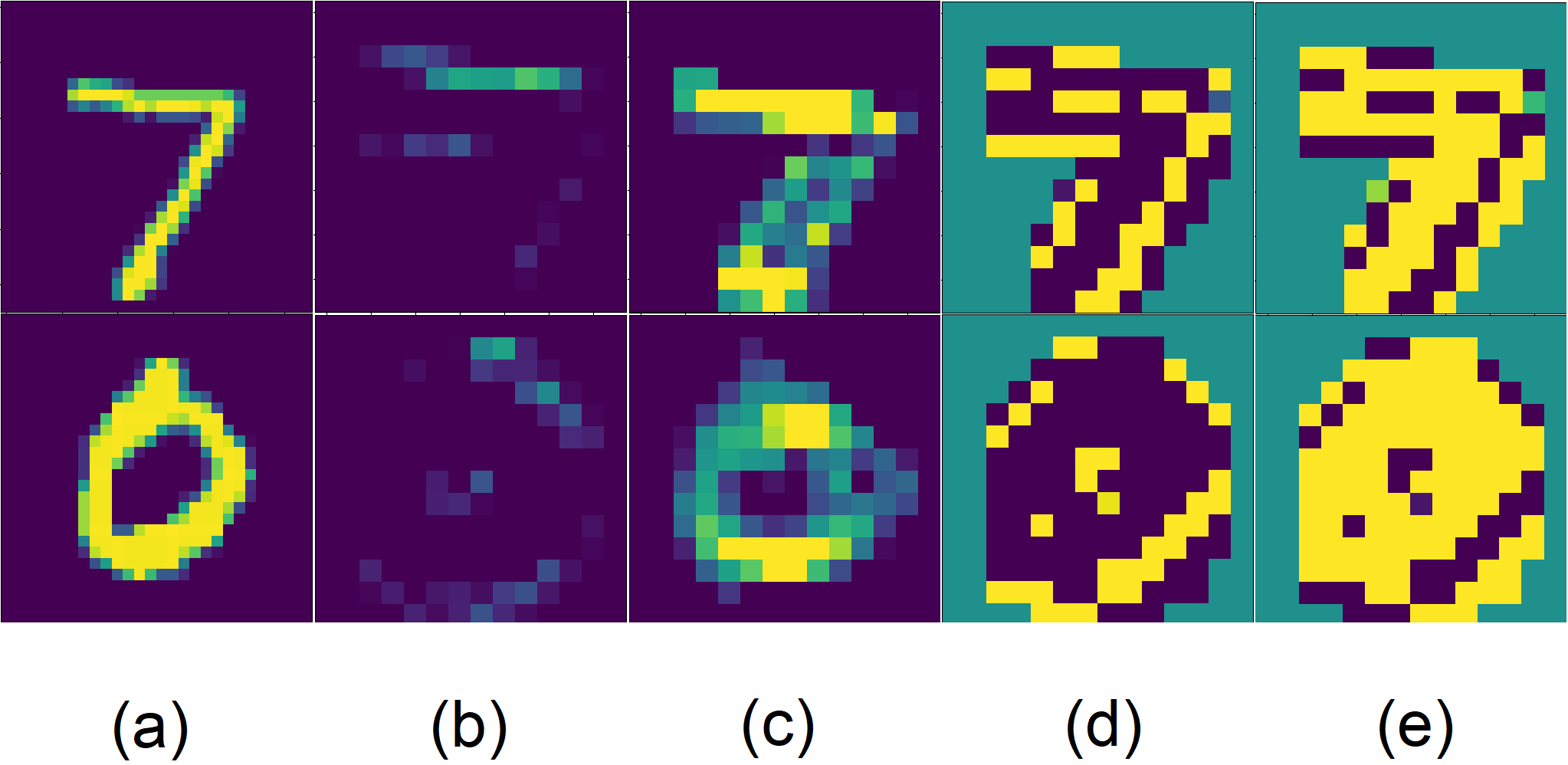}
\end{center}
   \caption{
   (a) MNIST images. (b) Feature maps of a trained ReLU ResNet18. (c) Same feature maps after teleportation. (d) Feature maps of a trained tanh ResNet18. (e) Same feature maps after teleportation.\vspace{-1.0cm}}
\label{fig:feature-maps}
\end{figure}

%C.f.~\cite{ArmentaJodoin20} for the proof that neural teleportation works for any activation function.

\subsection{Feature maps}

%Neural teleportation re-scales the neuron activation outputs with respect to the CoB~\cite{ArmentaJodoin20}. 
In order to put forward more concretely the effect that neural teleportation has on a neural network, we trained (on MNIST) and teleported two ResNet18 models: one with ReLU activations and one with tanh activations.  Feature maps of the original and teleported trained networks are shown in Fig.~\ref{fig:feature-maps}.  %A CoB-range of $0.9$ was used for both {\em inter-landscape} neural teleportations.  
As one can see, the feature maps before and after teleportation are, in both cases, very different.  This underlines the fact that while teleportation preserves the network function, it changes the features that the network has learned.

\section{Previous work}

%The theoretical work \cite{ArmentaJodoin20} shows that neural networks can be represented by the mathematical theory of quiver representations. However, optimization and experiments are not addressed and here we do exactly that.
%Concepts similar to that of neural teleportation have been proposed in the literature, the notion of positive scale invariance being the closest.
It was shown that positive scale invariance affects training by inducing symmetries on the weight space. As such, many methods have tried to take advantage of it, for example \cite{Badrinarayanan15a,Badrinarayanan15,Huang17,Neyshabur15}.  Our notion of teleportation gives a different perspective as i) it allows any non-zero real-value CoB to be used as scaling parameters, ii) it acts on any kind of activation functions, and iii) our approach do not impose any constraints on the structure of the network nor the data it processes. Also, neural teleportation does not require new update rules as it only has to be applied \textit{once} during training to produce an impact. % naturally adapts to existing gradient descent algorithms. 

%In \cite{Badrinarayanan15a} the authors equip the search space with a non-Euclidean metric to resolve the scale symmetries. This requires a new update rule for the weights during training. For teleportation one has to teleport the weights and the activations only once.

%In \cite{Huang17}, it is claimed that scaling-based symmetry in weight space produces the ill-conditioned problem, that can have a negative effect in the training of neural networks. They approach this problem with a new update rule. We show, however, that one can take advantage of the ill-condition problem by using teleportation instead.

\cite{Meng18,Neyshabur15} accounted for the fact that ReLU networks can be represented by the values of ``basis paths" connecting input neurons to output neurons. \cite{Meng18} made clear that these paths are interdependent and proposed an optimization based on it. They designed a space that manages these dependencies, proposing an update rule for the values of the paths that are later converted back into network weights. Unfortunately, their method only works for neural nets having the same number of neurons in each hidden layer. Furthermore, they guarantee no invariance across residual connections. %, so their algorithm has to be constrained to work only inside residual blocks and not across residual connections. 
This is unlike neural teleportation, which works for any network architecture, including residual networks.

Scale-invariance in networks with batch normalization has been observed by \cite{Arora19,Cho17}, but not in the sense of quiver representations.

%However, it is proved in \cite{ArmentaJodoin20} that there is invariance across residual connections, which can be seen when the quiver representation structure is made explicit. We allow our networks to be teleported across residual connections as well. 

Positive scale invariance of ReLU networks has also been used to prove that common flatness measures can be manipulated by the re-scaling factors \citep{Dinh17}. Here, we experimentally show that the loss landscape changes when teleporting a network with positive or negative CoB, regardless of its architecture and activation functions. Note that the proofs of \cite{Dinh17} are for two layer neural networks while we teleport deeper and much more sophisticated architectures, and provide empirical evidence of how teleportation sharpens the local loss landscape. %We don't claim anything about the flatness if the solution found by gradient descent versus gradient descent with teleportation.

\section{Neural teleportation and the loss landscape} 

Despite its apparent simplicity, neural teleportation has surprising consequences on the loss landscape.  In this section, we underline such consequences and lay empirical evidence for it.  

%, some being mathematical consequences of the work \cite{ArmentaJodoin20} and others not having been predicted.

\begin{figure}[tp]
\begin{center}
\includegraphics[scale=0.2]{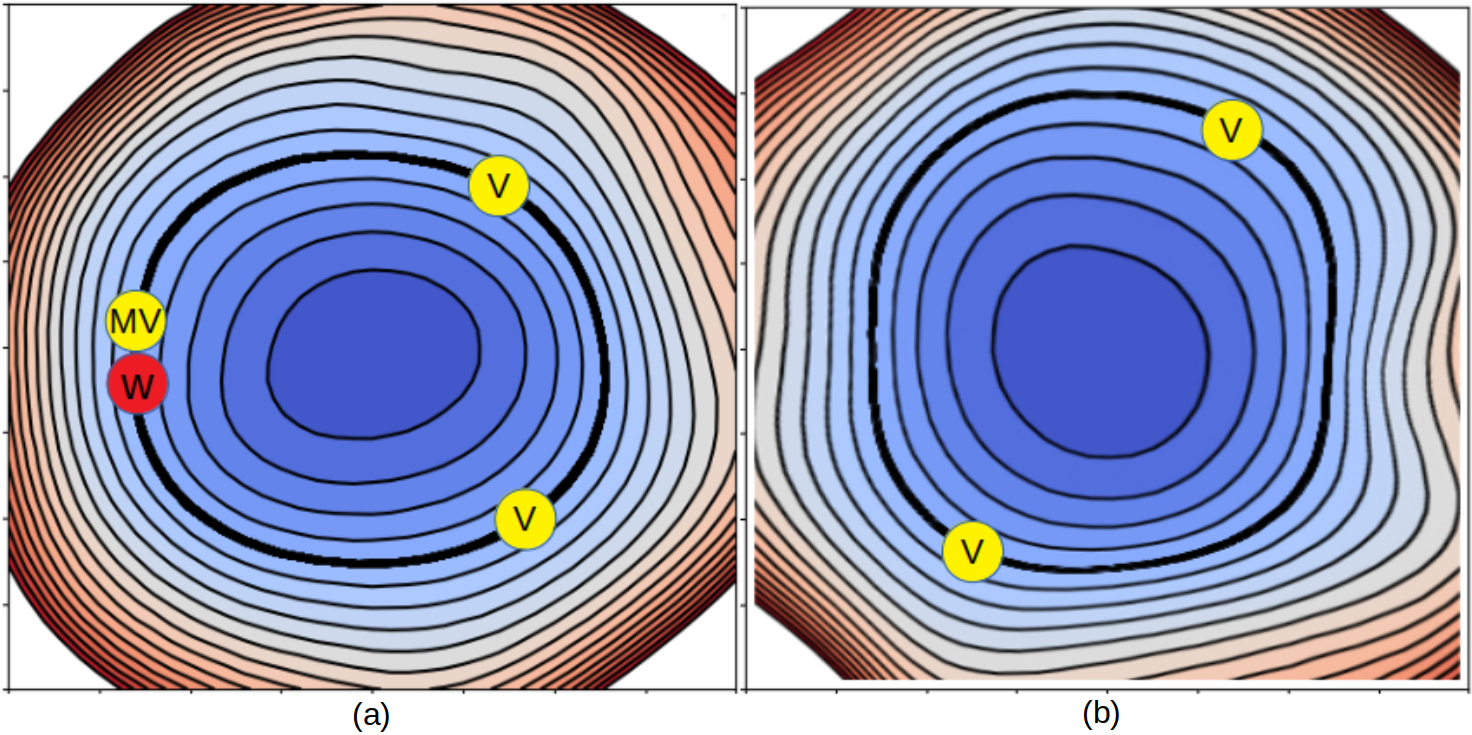}
\end{center}
   \caption{\vspace{-0.5cm}(a) 2D slice of a loss landscape with a W dot as the location of a given network.  The  V dots are two teleported versions of W inside the same landscape while the MV dot stands for a micro-teleportation of W.  (b) The dots are teleported versions of W in a different loss landscape.  Since neural teleportation preserves the network function, the six networks have rigorously the same loss value.\vspace{-0.75cm}}
   %[Left] 2D slice of the loss landscape of a ResNet18 with ReLU activation function for CIFAR-10.  The red \textbf{W} dot stands for the network weights after 25 epochs of training.  }
\label{fig:lossLandscape}
\end{figure}

\subsection{{\em Inter-} and {\em intra-}loss landscape teleportation }
\label{sec:interIntraTeleportation}

%Anything concerning optimization was not predicted by the results of \cite{ArmentaJodoin20}. In particular, the effects of neural teleportation on the loss landscape of neural networks.

As mentioned before, teleporting a positive scale invariant activation function $f_d$ with a positive $\tau_d$ results in $g_d=f_d$.  This means that the teleported network ends up inside the same loss landscape but with weights $V$ at a different location than $W$ (c.f., Fig.~\ref{fig:lossLandscape}(a)).  In other cases (for $\tau_d$ negative or non-positive scale invariant activation functions $f_d$) neural teleportation changes the activation function and thus the overall loss landscape.  For example, with $\tau_d<0$ and $f_d$ a ReLU function,  the teleportation of $f_d$ becomes: $g_d(x)=\tau_d \maxx(0, x/\tau_d) = \minn(0, \tau_d x/\tau_d) = \minn(0,x)$.%  We call this new activation function {\em NReLU} (Negative ReLU).

%Show that relu becomes nReLU (negative relu : min(0,x) )

Thus, the way CoB values are chosen has a concrete effect on how a network is teleported.  A trivial case is when  $\tau_\epsilon=1$ for every hidden neuron $\epsilon$, which leads to no transformation at all: $V=W$ and $g=f$.  For our method, we choose the CoB values by randomly sampling either of two distributions. The first one is a uniform distribution centered on $1$: $\tau_\epsilon \in [1-\sigma,1+\sigma]$ with $0<\sigma<1$.  We call $\sigma$ the {\em CoB-range}; the larger $\sigma$ is, the more different the teleported network weights $V$ will be from $W$.  Also, when this sampling operation is combined with positive scale invariant activation function (like ReLU), the teleported activation functions stay unchanged ($g=f$) and thus the new weights $V$ are guaranteed to stay within the same landscape as the original set of weights $W$ (as in Fig.~\ref{fig:lossLandscape} (a)). We thus call this operation an \textit{intra-landscape} neural teleportation. 

The other distribution is a mixture of two equal-sized uniform distributions: one centered at $+1$ and the other at $-1$:  $\tau_\epsilon \in [1-\sigma,1+\sigma] \cup [-1-\sigma,-1+\sigma].$  With high probability, a network teleported with this sampling will end up in a new loss landscape as illustrated in Fig.~\ref{fig:lossLandscape} (b).   We thus call this operation an \textit{inter-landscape} neural teleportation. 

%This is illustrated in Fig.~\ref{fig:lossLandscape}.  On the left is the 2D slice of a loss landscape computed following XYZ method~\cite{SOMETHING}, where we show teleportations produced by a within landscape sampling.  $W$ is the weight vector of a ResNet with ReLU activation function. On the right of Fig.~\ref{fig:lossLandscape}, we show a 2d slice of the loss landscape after teleporting a ResNet with a change landscape sampling.

\subsection{Loss level curves} 

Since $\tau_\epsilon$ can be assigned any non-zero real values,  a network can be teleported an infinite amount of times to an infinite amount of locations within the same landscape or across different landscapes.  This is illustrated in  Fig.~\ref{fig:lossLandscape} where $W$ is teleported to 5 different locations in two different loss landscapes.  Because of the very nature of neural teleportation, which preserves the network function, these 6 neural networks have the same loss value.  Thus, networks teleported in the same landscape {\em sit on the same level curve}.  

We validated this assertion by teleporting an MLP 100 times with an inter-landscape sampling and a CoB-range of $\sigma=0.9$.  While the mean average difference between the original weights $W$ and the teleported weights $V$ is of $0.08$ (a large value considering that the magnitude of $W$ is of $0.18$), the average loss difference was in the order of $10^{\mbox{-}10}$, {\em i.e.}, no more than a floating point error.

%As noticed in the proof of Theorem 4.9 in \cite{ArmentaJodoin20}, the neuron activations get scaled after teleportation, see Fig.~\ref{fig:feature-maps}.
\begin{figure}[tp]
\begin{center}
\includegraphics[scale=0.3]{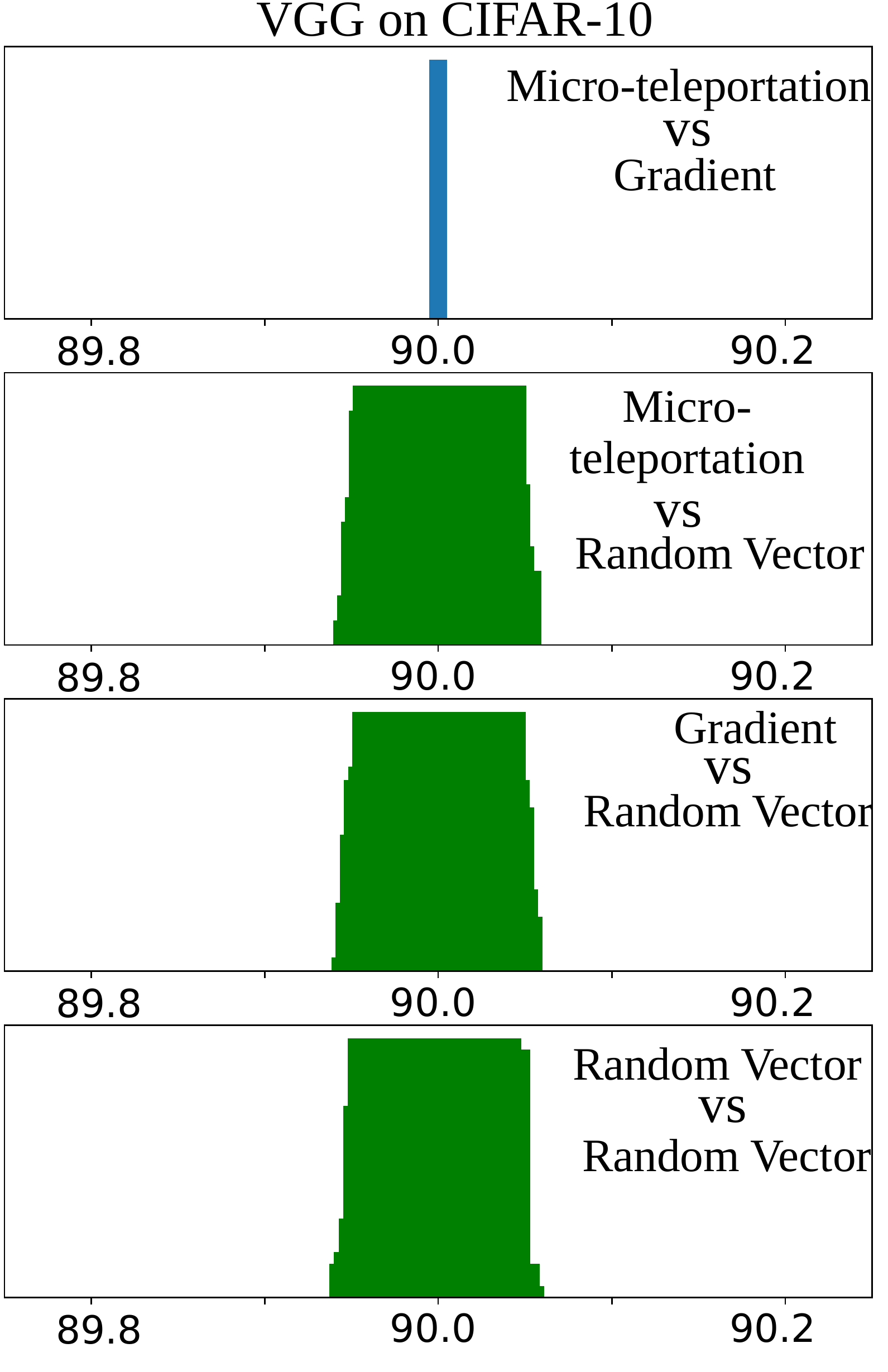}
\includegraphics[scale=0.3]{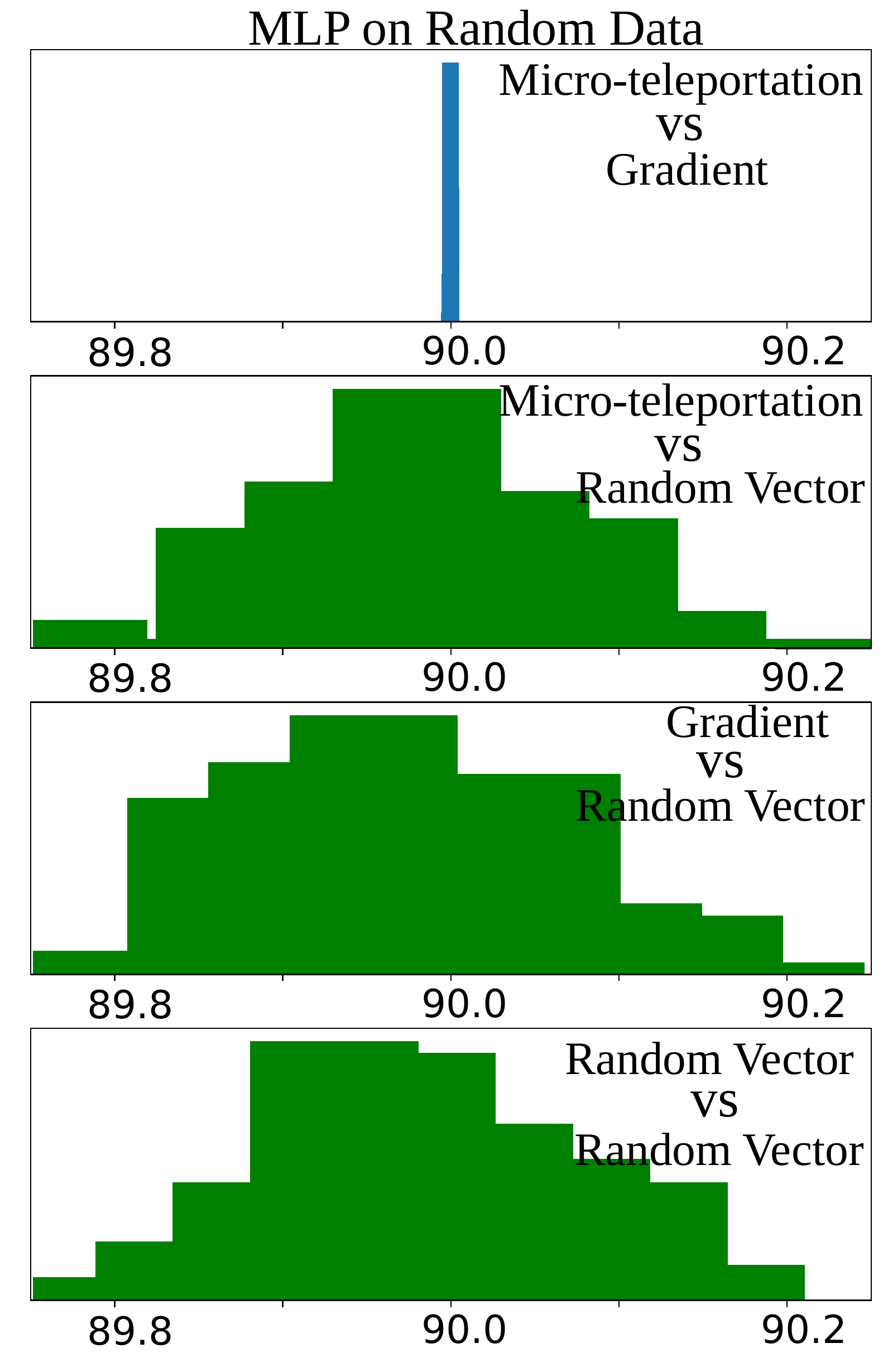}

\end{center}
   \caption{\vspace{-0.5cm}[Top] Angle histograms between micro-teleportation vectors and back-prop. gradients for VGGnet on CIFAR-10 data and MLP on random data.  The other rows are angle histograms  between micro-teleportation vectors and random vectors, between gradient and random vectors and between random vectors. \vspace{-0.5cm}
   }
\label{fig:micro1}
\end{figure} 

\subsection{Micro teleportation} 
One can easily see that the gradient of a function is always perpendicular to its local level curve (c.f. \cite{Spivak} chap. 2). However,  back-propagation computes noisy gradients that depend on the amount of data used, that is, the batch-size. Thus, the noisy back-propagated gradients do not {\em a priori}  have to be strictly perpendicular to the local loss level curves.  %This may give rise to the legitimate concern of whether teleported networks have more erratic, and thus error prone, gradients.

This concern can be (partly) answered via the notion of \textit{micro-teleportation}, which derives from the previous subsection.  Let's consider the intra-landscape teleportation of a network with positive scale invariant activation functions (like ReLU) with a CoB-range $\sigma$ close to zero.  In that case, $\tau_\epsilon\approx 1$ for every neuron $\epsilon$ and the teleported weights $V$ (computed following Eq.(\ref{eq:neuralTeleport})) end up being very close to $W$.  We call this a \textit{micro-teleportation} and illustrate it in Fig.~\ref{eq:neuralTeleport} (a) (the MV dot illustrates the micro teleportation of W).

Because $V$ and $W$ are isomorphic, they both lie on the same loss level curve. % While $V$ and $W$ are similar sets of weights, from the previous subsection, they both lie on the same loss isoline.
Thus, if $\sigma$ is small enough, the vector $\overrightarrow{WV}$ between $W$ and $V$ is locally co-linear to the local loss level curve.  We call  $\overrightarrow{WV}$ a {\em micro-teleportation vector}.

A rather counter-intuitive empirical property of micro-teleportations is that $\overrightarrow{WV}$ is perpendicular to {\em any back-propagated gradient} across different batch sizes because we know that the back-propagated gradient depends on the batch size but the teleportation doesn't.  % as $\overrightarrow{WV}$ is not computed with a set of data nor a loss function.
%Since a gradient is by definition perpendicular to the local level curve (c.f. \cite{Spivak} chap. 2) a rather counter-intuitive consequence of micro-teleportation is that $\overrightarrow{WV}$ must be perpendicular to any back-propagated gradient. % This can be appreciated in Fig.~\ref{fig:lossLandscape}, where the vector $\overrightarrow{WV}$ becomes perpendicular to the gradient when $\tau \to 1$. 
 This surprising observation leads to the following conjecture.
%any micro-teleportation vector must be perpendicular to {\em any} back-propagated gradient, generated from any set of data and any loss function. 

\begin{Conjecture} \label{conjecture:micro}
    For any neural network, any dataset and any loss function, there exists a sufficiently small CoB-range $\sigma$ so that every micro-teleportation produced with it is perpendicular to the back-propagated gradient with any batch size.
    %For any neural network, any dataset and any loss function, there exists a micro-teleportation vector that is perpendicular to the back-propagated gradient.
\end{Conjecture}

%This conjecture derives from the fact that $\overrightarrow{WV}$ is not computed with a set of data nor a loss function.  %, $\overrightarrow{WV}$ should be perpendicular to back-propagated gradients regardless of the loss function and the data used to compute it.
We empirically assessed this conjecture by computing the angle between micro-teleportation vectors and back-propagated gradients of four models (MLP, VGG, ResNet and DenseNet) on three datasets (CIFAR-10, CIFAR-100 and random data) $2$ different batch sizes (8 and 64) with a CoB-range $\sigma=0.001$. A cross-entropy loss was used for all models.  Fig.~\ref{fig:micro1} shows angular histograms for VGG on CIFAR-10 and an MLP (with one hidden layer of 128 neurons) on random data (results for other configurations are in the appendix). The first row shows the angular histogram between micro-teleportation vectors and gradients.  We used batch sizes of 8 for MLP and 64 for VGG.  As can be seen, both histograms exhibit a clear Dirac delta on the $90^{\mbox{\tiny o}}$ angle.  As a mean of comparison, we report angular histograms between micro-teleportation vectors and random vectors, between back-propagated gradients and random vectors, and between random vectors.  While random vectors in a high-dimensional space are known to be quasi-orthogonal (c.f work of \cite{Kainen2020}), by no means are they exactly orthogonal, as shown by the green histograms. The Dirac delta of the first row versus the wider green  distributions is a clear illustration of our conjecture.

These empirical findings suggest that although back-propagation computes noisy gradients, their orientation in the weight space does not entirely depend on the data nor the loss function. In other words, back-propagation computes gradients that point to directions perpendicular to micro-teleportation vectors, but micro-teleportation vectors do not depend on the loss nor the data. Finally, we note that this conjecture is false if one adds a regularization term that does not depend directly on the output of the network, for example, $l2$ regularization adds a term that depends on the weights but not on the output of the network.

%These empirical evidence underlines that back-propagation is always choosing a direction perpendicular to micro-teleportation vectors.

\begin{figure}[tp]
\begin{center}
\includegraphics[width=1\linewidth,height=3.5cm]{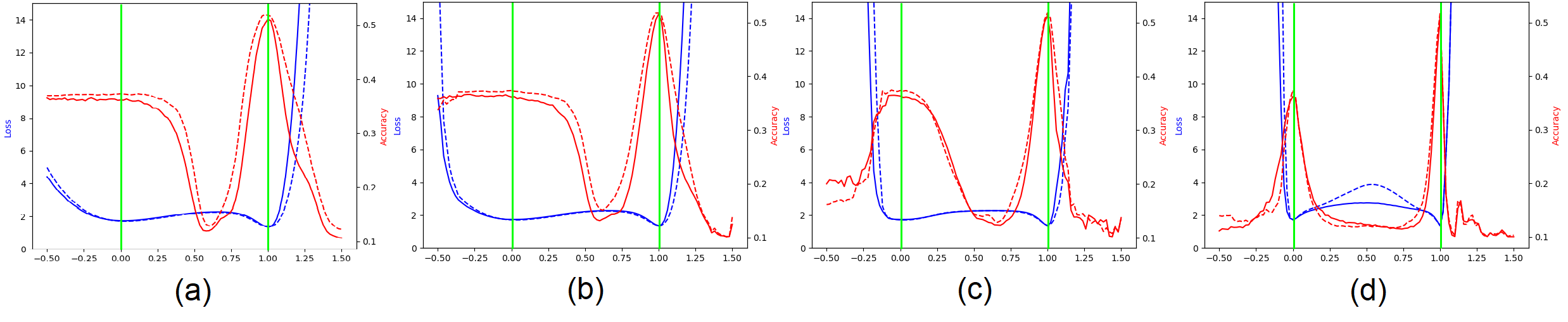}
\end{center}
   \caption{\vspace{-0.5cm}(a) Loss/accuracy profiles obtained by linearly interpolating between two optimized MLP $A$ and $B$. Network $A$ is for $x=0$ and $B$ for $x=1$ (green vertical lines). Dotted lines are for training and solid lines for validation. Remaining plots are similar interpolations but between teleported versions of $A$ and $B$ with CoB range $\sigma$ of (b) $0.6$, (c) $0.9$, and (d) $0.99$.\vspace{-0.5cm}}
\label{fig:interpol1}
\end{figure}

\subsection{Teleportation and landscape flatness}

It has been shown by \cite{Dinh17} for positive scale invariant activation functions, that one can find a CoB (called {\em reparametrization} in their paper) so that the most commonly used measures for flatness can be manipulated to obtain a sharper minimum with the exact same network function. We empirically show that  neural teleportation  systematically sharpens the found minima, independently of the architecture or the activation functions.

A commonly used method to compare the flatness of two trained networks is to plot the 1D loss/accuracy curves on the interpolation between the two sets of weights. It is also well known that small batch sizes produce flatter minima than bigger batch sizes. We trained on CIFAR-10 a 5 hidden-layer MLP two times, first with a batch size of $8$ (network $A$) than with a batch size of $1024$ (network $B$). Then, as done by \cite{Li18}, we plotted the 1D loss/accuracy curves on the interpolation between the two weight vectors of the networks (c.f. Fig.~\ref{fig:interpol1}(a)).   We then performed the same interpolation but between the teleportation of A and B with CoB-ranges $\sigma$ of $0.6$, $0.9$, and $0.99$. As can be seen from  Fig.~\ref{fig:interpol1}, the landscape becomes sharper as the CoB-range increases.  Said otherwise, a larger teleportation leads to a locally-sharper landscape.  More experiments with other models can be found in the appendix.

%To summarise, in order to teleport a neural network we choose a CoB that preserves the architecture of the network, then apply the CoB to the weights and to the activation functions as mentioned above.

%We note that the trivial CoB given by assigning the number 1 to every neuron of the network, produces the same network after teleportation. Also, if the chosen CoB is given by 1's and -1's, then teleporting twice will produce the same network again.

%If the CoB range is greater than 1, the within landscape sampling may produce a CoB with negative values, and therefore the sampling will pass through a discontinuity of the action of the CoB group on neural networks. So every one of our experiments uses a CoB range strictly less than 1. 

%\begin{figure*}
%\begin{center}
%\end{center}
%   \caption{We take networks $A$ and $B$ from Fig.~\ref{fig:interpol1}. The first plot shows the interpolation between a teleportation of $A$ and a teleportation of $B$, both with the same change of basis randomly chosen from a within landscape sampling. The second plot is the interpolation between another teleportation of $A$ and a teleportation of $B$ with the same change of basis (within landscape sampling) but different than that of the first plot. The third plot is the interpolation between a teleportation of $A$ and a teleportation of $B$ with the same change of basis sampled with change landscape sampling.}
%\label{fig:interpol2}
%\end{figure*}

\begin{figure}[tp]
\begin{center}
\includegraphics[width=.24\linewidth]{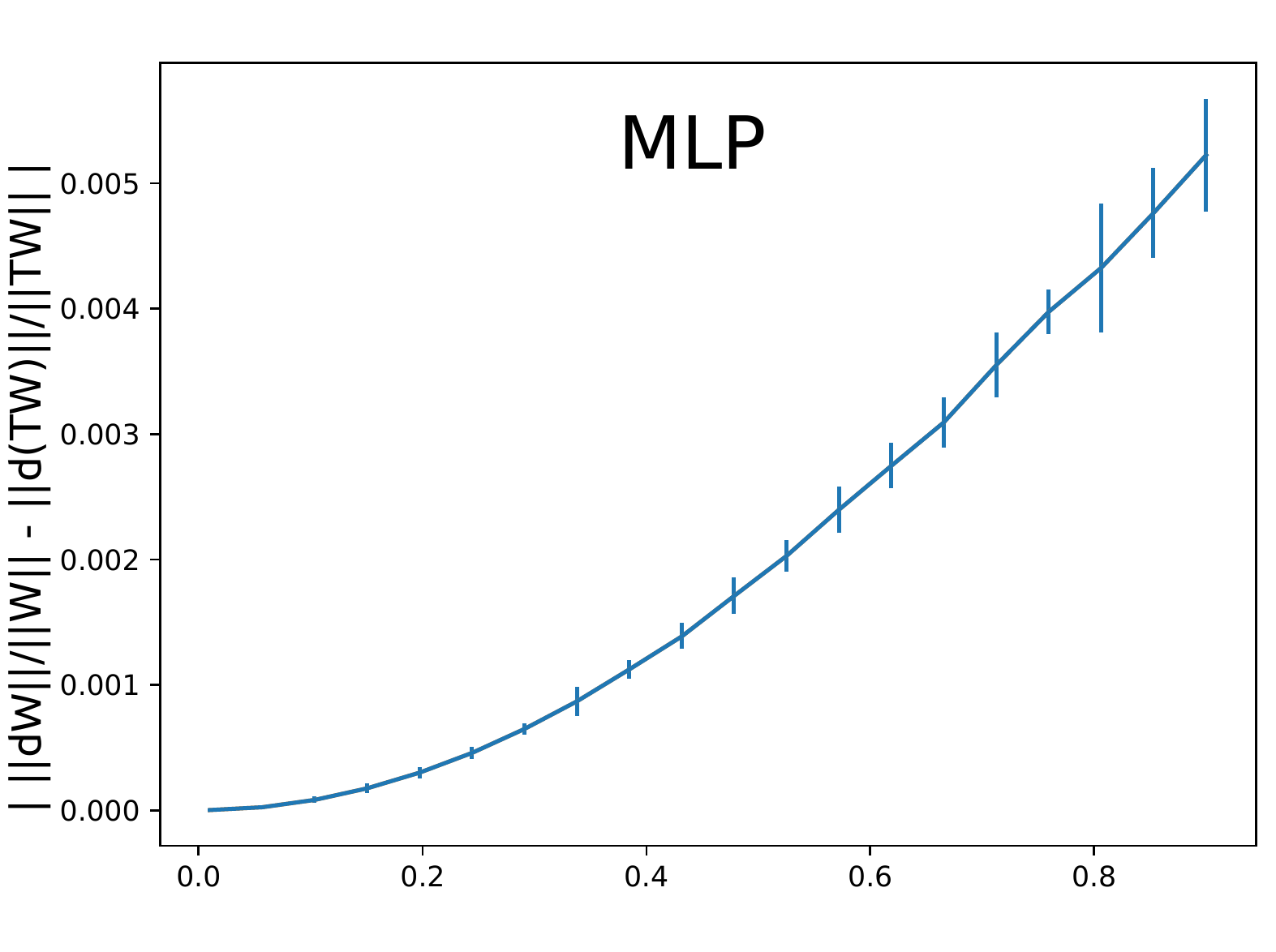}
\includegraphics[width=.24\linewidth]{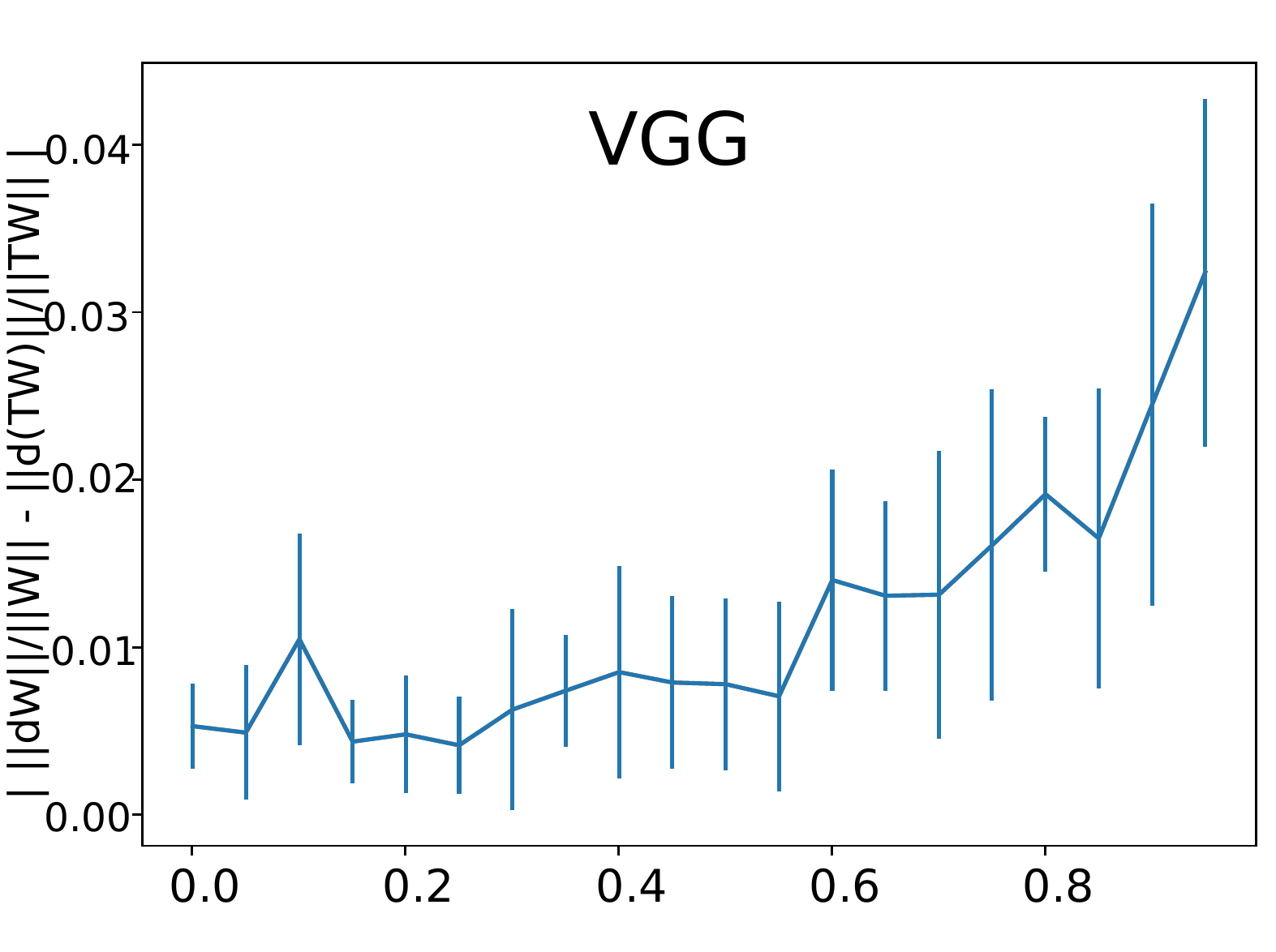}
\includegraphics[width=.24\linewidth]{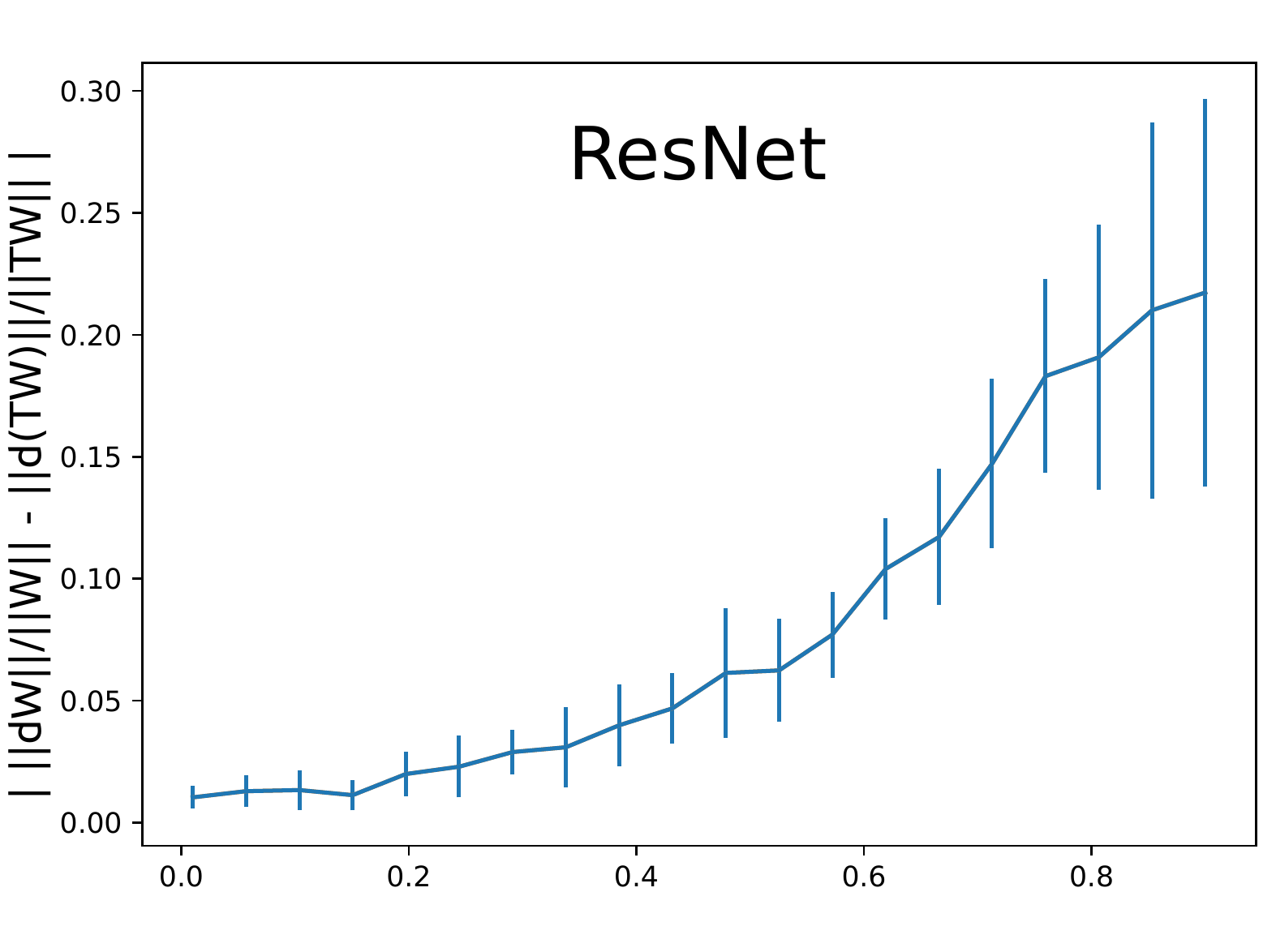}
\includegraphics[width=.24\linewidth]{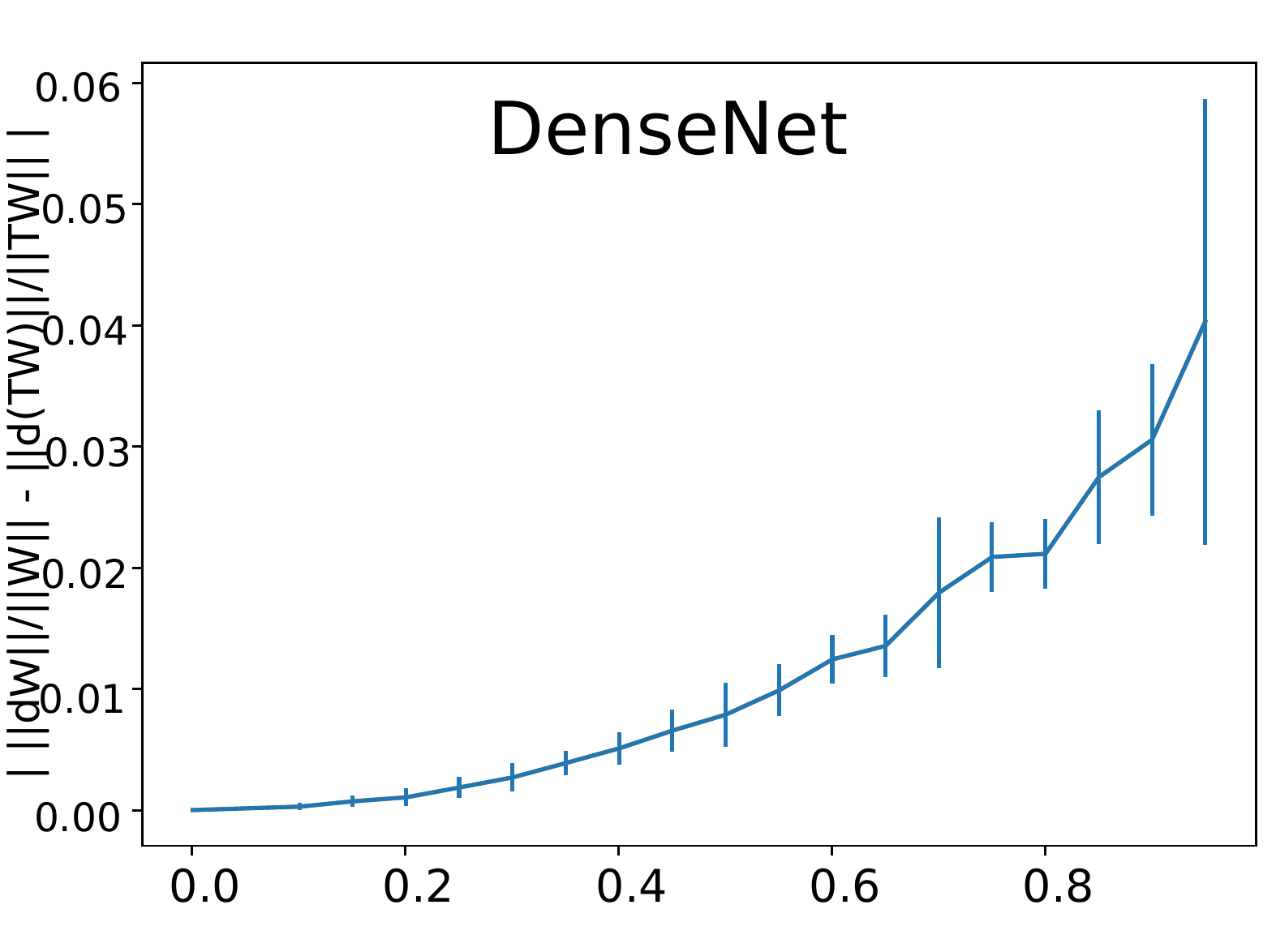}
\end{center}
   \caption{\vspace{-0.5cm}Mean absolute difference ($\pm$ std-dev) between the back-propagated gradients' magnitudes of teleported networks and their original non-teleported network.  Larger CoB generate larger gradients.  Results are for CIFAR-10. Plots for these same models on CIFAR-100 can be found in the appendix.\vspace{-0.5cm}% The $y$ axis is the absolute value of the difference between the norms of the ratio of the norms of the gradient vector to weight vector between a network and a teleportation with change of basis range measured by the $x$ axis in the interval $(0.1,0.9)$. We computed the mean curve and standard deviation bars over 20 different teleportations. 
   } \label{fig:grad_change}
\end{figure}

\vspace{-0.2cm}
\section{Optimization}
\vspace{-0.2cm}
In the previous section, we showed that teleportation moves a network along a loss level curve and sharpens the local loss landscape.  In this section, we show that teleportation increases the magnitude of the local normalized gradient and accelerates the convergence rate when used during training, even only \textit{once}. Moreover, we empirically show that this impact is controlled by the CoB-range factor $\sigma$.

We remark that once a neural network is initialized it is assigned a specific network function and, in principle, a teleported network of this initialized network (which will have the same network function) doesn't have to train better than the original network and, nevertheless, the teleported network trains better.

\begin{figure*}
    \begin{center}
    \includegraphics[width=0.40\linewidth]{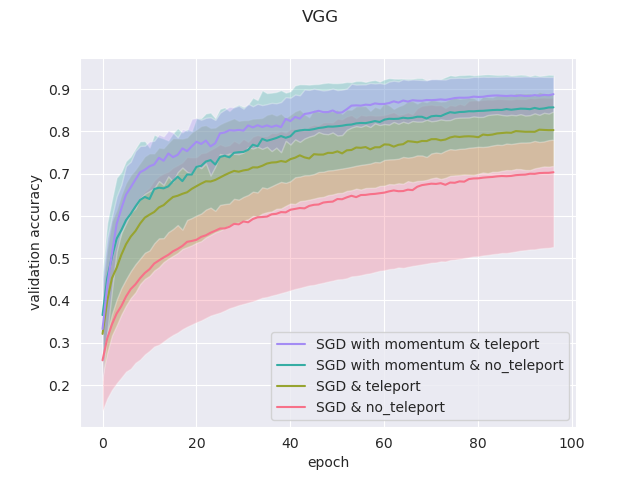}
    \includegraphics[width=0.40\linewidth]{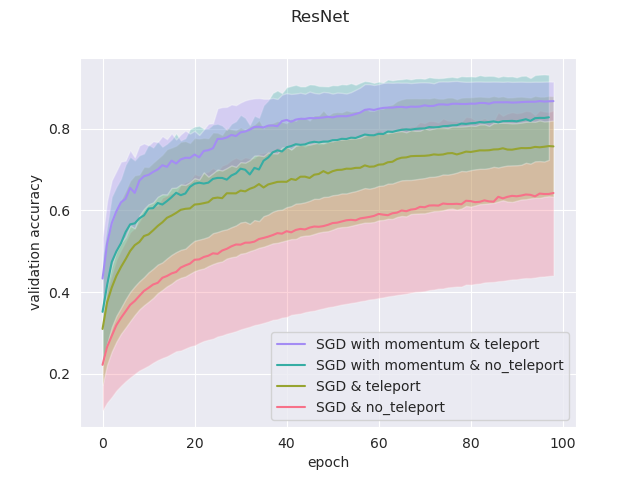}
    \includegraphics[width=0.40\linewidth]{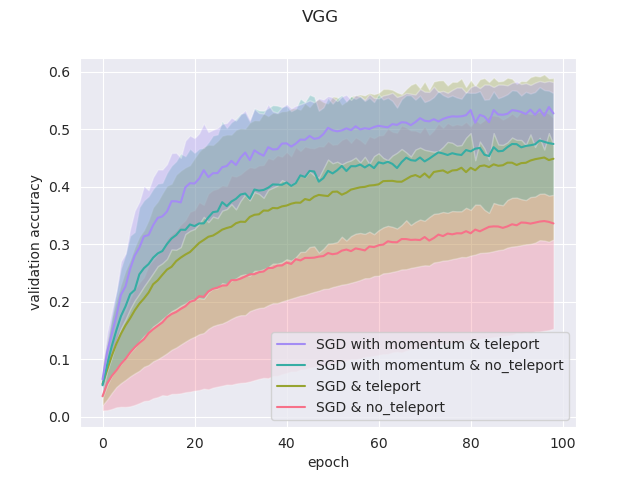}
    \includegraphics[width=0.40\linewidth]{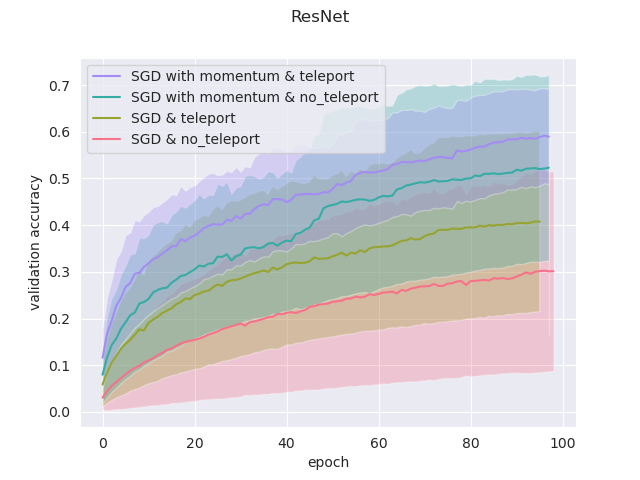}

    \end{center}
       \caption{\vspace{-0.75cm}Validation accuracies for VGG and ResNet on CIFAR-10 (top row) and CIFAR-100 (bottom row). %We used weight decay of 0.01. 
       The curves are produced by averaging over $5$ runs for all learning rates ($0.01$, $0.001$ and $0.0001$) with shades representing $\pm$  std. dev.\vspace{-0.75cm}}
    \label{fig:training_curves}
\end{figure*}

\subsection{Back-propagated gradients of a teleportation}

%\section{The gradient of a teleportation} \label{sec:telep_grad}

It has already been noticed by \cite{Dinh17,Neyshabur15} that under positive scaling of ReLU networks, the gradient scales inversely than the weights with respect to the CoB. Here, we show that the back-propagated gradient of a teleported network has the same property, regardless of the architecture of the network, the data, the loss function and the activation functions.

Let $(W,f)$ be a neural network with a set of weights $W$ and activation functions $f$. We denote  $W^{\left[ \ell \right]}$ the weight tensor of the $\ell$-th layer of the network, and analogously for the teleportation $V^{\left[ \ell \right]}$. The CoB $\tau$ at layer $\ell$ is denoted $\tau^{\left[ \ell \right]}$, which is a column vector of non-zero real numbers. Let's also consider a data sample $(\mathbf{x},t)$ with $\mathbf{x}$ the input data and $t$ the target value and $dW,dV$ the gradient of the networks $(W,f)$ and $(V,g)$ with respect to $(\mathbf{x},t)$.  Following  Eq.~(\ref{eq:neuralTeleport}), we have that
\begin{eqnarray}
\label{eq:re-scale-weight}
    V^{\left[ \ell \right]} = \frac{1}{\tau^{\left[ \ell -1 \right]}} \bullet W^{\left[ \ell \right]} \bullet \tau^{\left[ \ell \right]},
\end{eqnarray}

\begin{figure*}[thp]
    \begin{center}
    \includegraphics[width=0.40\linewidth]{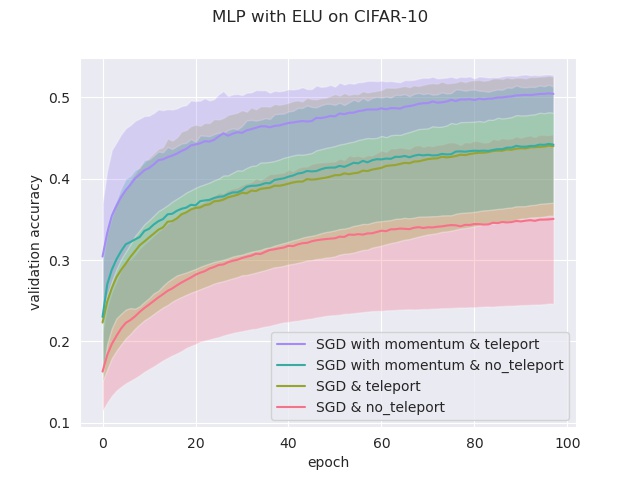}
    \includegraphics[width=0.40\linewidth]{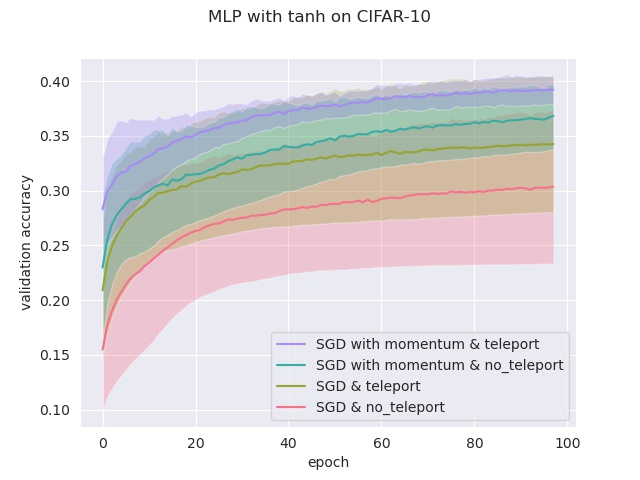}
    \end{center}
       \caption{\vspace{-0.5cm}Validation plots produced over $5$ runs over three learning rates ($0.01$, $0.001$ and $0.0001$) of MLPs with activations ELU [left] and Tanh [right] on CIFAR-10.\vspace{-0.5cm}}
    \label{fig:training_curves_otheracts}
\end{figure*}

where the operation $\frac{1}{\tau^{\left[ \ell -1 \right]}} \bullet  W^{\left[ \ell \right]}$ multiplies the columns of the matrix $W^{\left[ \ell \right]}$ by the coordinate values of vector $\frac{1}{\tau^{\left[ \ell -1 \right]}}$, while the operation $W^{\left[ \ell \right]} \bullet \tau^{\left[ \ell \right]}$ multiplies the rows of matrix $W^{\left[ \ell \right]}$ by the coordinate values of the vector $ \tau^{\left[ \ell \right]}$.

\begin{Theorem} \label{thm:grad_tel}
    Let $(V,g)$ be a teleportation of the neural network $(W,f)$ with respect to the CoB $\tau$. Then 
    \[
        dV^{\left[ \ell \right]} = \tau^{\left[ \ell - 1 \right]} \bullet  dW^{\left[ \ell \right]} \bullet \frac{1}{\tau^{\left[ \ell \right]}} 
   \]
    for every layer $\ell$ of the network (proof in the appendix). 
\end{Theorem}

If we look at the magnitude of the teleported gradient, we have that
\[
    \| dV \| = \sqrt{ {\displaystyle\sum}_{i,j} \left(dW_{i,j} \frac{\tau_j}{\tau_i} \right)^2 }. %(w_1 \tau_1)^2+(w_2 \frac{\tau_2}{\tau_1})^2+ \cdots + (w_{n-1}\frac{\tau_{n-1}}{\tau_n})^2 + (w_n \frac{1}{\tau_n})^2}
\]
We can see that the ratio $\tau_a^2/\tau_b^2$ appears multiplying the squared non-teleported gradient $dW^2$. For an {\em intra-}landscape teleportation, $\tau_a$ is randomly sampled from a uniform distribution $[1-\sigma,1+\sigma]$ for $0<\sigma<1$. Since $\tau_a,\tau_b$ are independent random variables, the mathematical expectation of this squared ratio is
%\[
%    \mathbb{E} \left[ \tau_a^2/\tau_b^2 \right] = \mathbb{E} \left[ \tau_a^2 \right] \cdot \mathbb{E} \left[ 1/\tau_b^2 \right].
%\]
%Then, after a few computations we obtain
\[
    \begin{array}{lcl}
    \mathbb{E} \left[ \tau_a^2/\tau_b^2 \right] &= &\mathbb{E} \left[ \tau_a^2 \right] \cdot \mathbb{E} \left[ 1/\tau_b^2 \right]\\
        %\mathbb{E} \left[ \tau_a^2/\tau_b^2 \right]  & = & {\displaystyle\int_{[1-\sigma,1+\sigma]^2}} \tau_a^2/\tau_b^2 P(\tau_a) P(\tau_b)  \\
        & = & \int_{1-\sigma}^{1+\sigma} \tau_a^2 P(\tau_a)d\tau_a  \cdot \int_{1-\sigma}^{1+\sigma} P(\tau_b)/\tau_b^2 d\tau_b \\
%        & = & \left( \frac{1}{3} (\sigma^2+3) \right) \left( - \frac{1}{\sigma^2-1} \right) \\
        & = & \frac{ \sigma^2+3}{3(1-\sigma^2)}.
    \end{array} 
\]
Thus, when $\sigma\rightarrow 0$ ({\em i.e.}, no teleportation as described in section~\ref{sec:interIntraTeleportation}), then $\mathbb{E} \left[ \tau_a^2/\tau_b^2 \right] \rightarrow 1$ which means that on average the gradients are multiplied by $1$ and thus remain unchanged. But when $\sigma \to 1$, then $\mathbb{E} \left[ \tau_a^2/\tau_b^2 \right] \to \infty$ which means that the gradients magnitude gets multiplied by an increasingly large factor. 

We empirically validate this proposition with four different networks in Fig.~\ref{fig:grad_change} (plots for the other models and datasets can be found in the appendix).  There we put the CoB-range $\sigma$ on the x-axis versus 20 different teleportations for which we computed the absolute difference of normalized gradient magnitude: $\left| \| dW \| / \| W \| - \|dV\|/\|V\| \right|$. We can see that a larger CoB-range leads to a larger difference between the normalized gradient magnitudes.  

This shows that randomly teleporting a neural network increases the sharpness of the surrounding landscape and thus the magnitude of the local gradient.  This holds true for any network architecture and any dataset. This analysis also holds true for {\em inter-}landscape CoB sampling since the CoB-range $\sigma$ appears squared always.

\begin{figure*}[tp]
\begin{center}
\includegraphics[width=.48\linewidth]{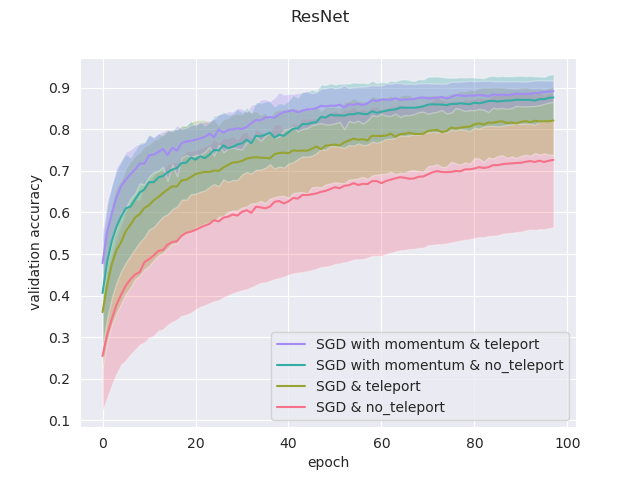}
\includegraphics[width=.48\linewidth]{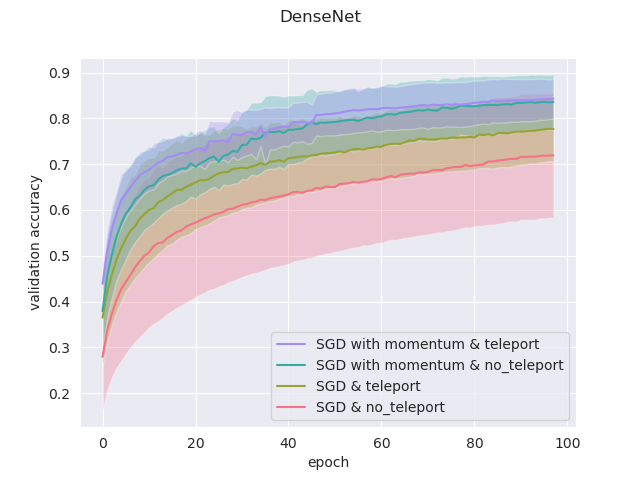}

\end{center}
   \caption{\vspace{-0.5cm}Validation plots produced over $5$ runs over three learning rates ($0.01$, $0.001$ and $0.0001$) for ResNet and DenseNet with normal initializations on CIFAR-10. \vspace{-0.75cm}}
\label{fig:init}
\end{figure*}

\subsection{Gradient descent and teleportation} \label{sec:optimization}
%\fi
%According to previous works \cite{Meng18,Neyshabur15}, the symmetries in weight space provided by positive scale invariance improperly influence optimization. We empirically show, however, that by choosing the COB-range from the algebraic action on neural networks given by the CoB group, one can improve optimization across different models, datasets, optimizers, activation functions and initializations.

%An immediate consequence of the previous proposition is that teleporting a network just after its initialization sets it in an area of the landscape where the slope is steeper and thus that convergence should be quicker.

From the previous subsection we obtain that the CoB-range controls the difference on the normalized gradients. Here we empirically show that the CoB-range also has an influence on the training when teleportation is applied \textit{once} during training.

In order to assess this, we trained four models: an MLP  with 5 hidden layers with 500 neurons each, a VGGnet, a ResNet18, and a DenseNet, all with ReLU activation.  Training was done on two datasets (CIFAR-10 and CIFAR-100) with two optimizers (vanilla SGD and SGD with momentum), and three different learning rates (0.01, 0.001 and 0.0001) for a total of 72 configurations.  For each configuration, we trained the network with and without neural teleportation right after initialization. Training was done five times for 100 epochs.  The chosen CoB range $\sigma$ is $0.9$ with an {\em inter}-landscape sampling for all these experiments. The teleported and non-teleported networks were initialized with the same randomized operator following the ``Kaiming" method.

\begin{figure*}[tp]
\begin{center}
\includegraphics[width=0.48\linewidth]{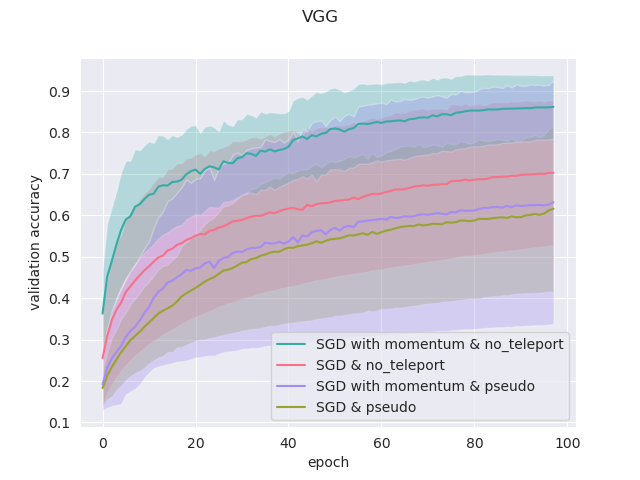}
\includegraphics[width=0.48\linewidth]{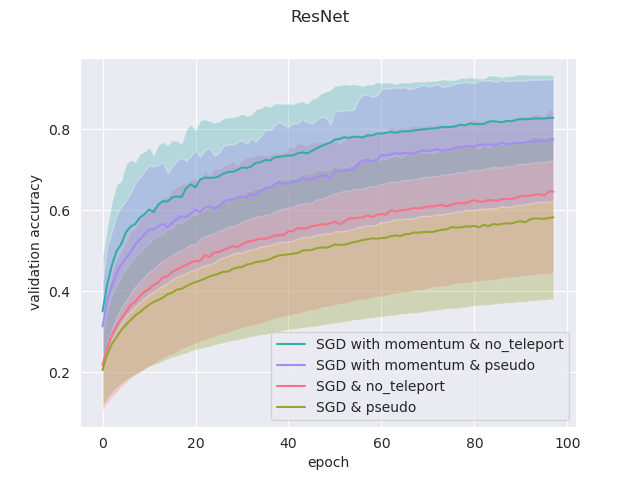}
\end{center}
   \caption{\vspace{-0.5cm}Validation plots produced over $5$ runs over three learning rates ($0.01$, $0.001$ and $0.0001$) for  VGG and ResNet on CIFAR-10 comparing a usual training and a training with pseudo-teleportation.\vspace{-0.75cm}} \label{fig:pseudo}
\end{figure*}

\begin{figure}[h]
\begin{center}
 %   \begin{subfigure}
   \includegraphics[scale=0.35]{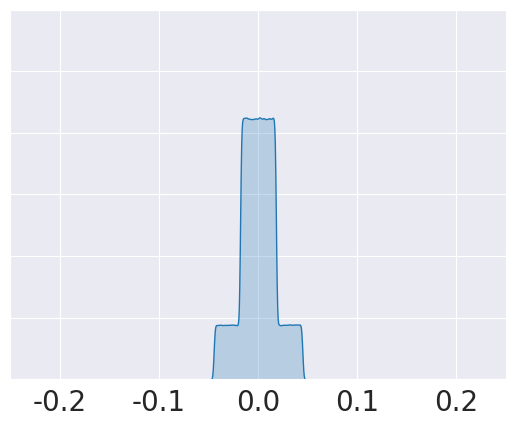}
%\end{subfigure}
%\begin{subfigure}
   \includegraphics[scale=0.35]{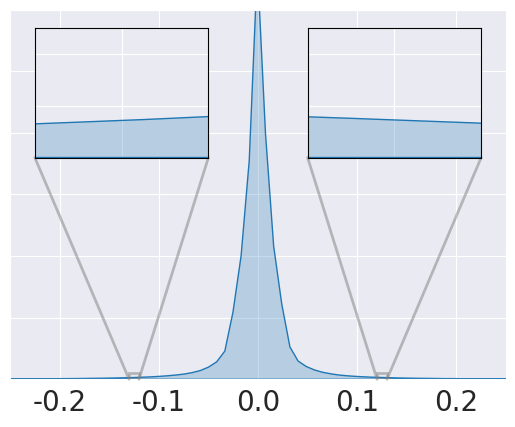}
%\end{subfigure}
\end{center}
   \caption{\vspace{-0.5cm}Weight histogram of an MLP before teleportation [Left] and after teleportation [Right] with CoB-range of $0.9$. Initialization is the standard on PyTorch which is uniform and contains values between $-.05$ and $.05$ while the teleported weights are more broadly distributed as shown in the zooms.\vspace{-0.5cm}}  %More histograms are in the supp. material.
   \label{fig:hist_weights_teleport}
\end{figure}

This resulted into a total of 720 training curves that we averaged across the learning rates ($\pm$ std-dev shades), see Fig.~\ref{fig:training_curves} (plots for the other models and datasets can be found in the appendix).  
%The light green and red boxes are results from normal training while the darker green and red boxes are results from the same training but with neural teleportation after initialization.  
As can be seen, neural teleportation accelerates gradient descent (with and without momentum) for every model on every dataset.

To make sure that these results are not unique to ReLU networks, we trained the MLP on CIFAR-10 and CIFAR-100 with three different activation functions : LeakyReLU, Tanh and ELU, again with and without neural teleportation after the uniform initialization, which is the default PyTorch init mode for fully connected layers.  As can be seen from Fig.~\ref{fig:training_curves_otheracts} (plots for the other models and datasets can be found in the appendix), here again \textit{one} teleportation accelerates gradient descent (with and without momentum) across datasets and models.

In order to measure the impact of the initialization procedure, we ran a similar experiment with a basic Gaussian initialization as well as an Xavier initialization. We obtained the same pattern for the Gaussian case, and for the Xavier initialization teleportation helps the training of the VGG net, while for the others the difference is less prominent, see Fig.~\ref{fig:init} (plots for the other models and datasets can be found in the appendix).% and still obtained the same pattern. Namely, teleportation at initialization accelerates training independently of the initialization method. This can be appreciated in Fig.~\ref{fig:init}.

%As can be seen, the weight distributions are significantly different.  We show in Fig.~\ref{fig:same_distr} that reproducing the same distribution on the weights does not leads to a better acceleration on the training than teleportation. 

%In order to show that convergence speed up does not depend on the weight distribution but on the teleportation process itself, we randomly generated weights so their distribution follows that of Figure~\ref{fig:hist_weights_teleport}~[right].  

Given a neural network $W$ we produced another one $V$ by first teleporting $W$ to $\tau W$ and then sampling $V$ from the sphere with center $W$ and radius vector $W - \tau W$. We called this a \textit{pseudo-teleportation}.  We then trained the four models with the same regime, comparing pseudo-teleportation to no teleportation. Results in Fig.~\ref{fig:pseudo} reveal that pseudo-teleportation does not improve gradient descent training (plots for the other models and datasets can be found in the appendix).

We show in Fig.~\ref{fig:hist_weights_teleport} the weight histograms of an MLP network initialized with a uniform distribution (PyTorch's default) before (left) and after (right) teleportation.  

\begin{figure}[tp]
\begin{center}
\includegraphics[scale=0.3]{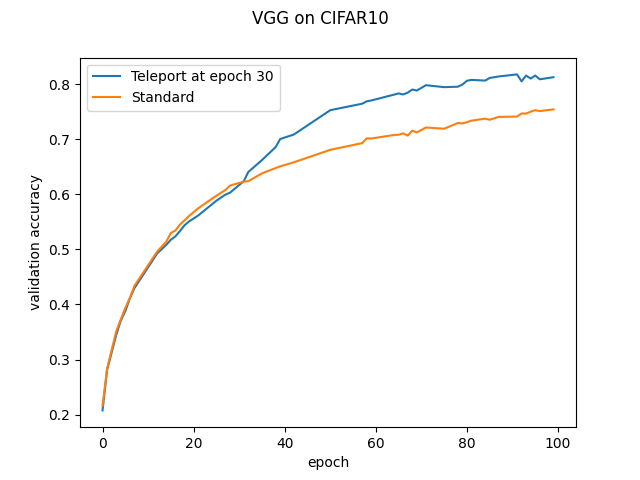}
\includegraphics[scale=0.3]{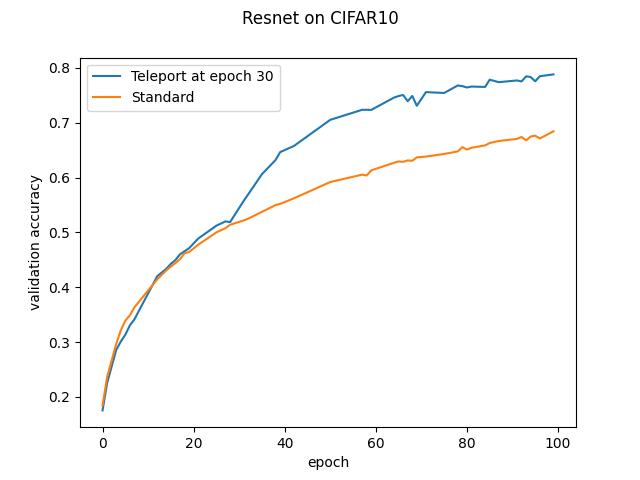}
\end{center}

   \caption{\vspace{-0.5cm}Validation accuracy for VGG [Left] and ResNet[Right]. We plot two curves for each model, the orange curves are the usual SGD training and the blue curves are SGD with one teleportation at epoch 30.\vspace{-0.75cm}}  %More histograms are in the supp. material.
   \label{fig:jump}
\end{figure}

\begin{figure}[tp]
\begin{center}
\includegraphics[scale=0.3]{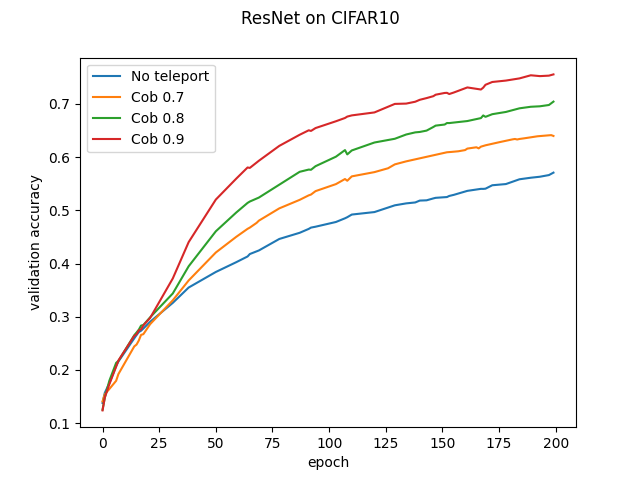}
\includegraphics[scale=0.3]{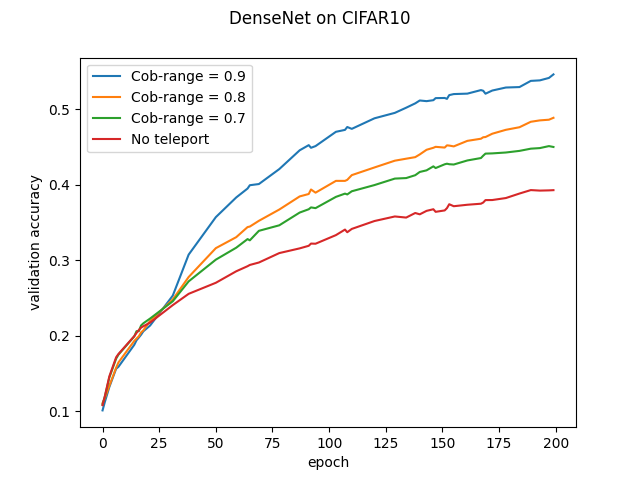}

\end{center}

   \caption{\vspace{-0.5cm}Validation accuracy for ResNet [Left] and DenseNet[Right]. We plot four curves for each model each one corresponding to applying one teleportation at epoch 30 with different CoB-ranges.\vspace{-0.75cm}}  %More histograms are in the supp. material.
   \label{fig:jump_diffcobs}
\end{figure}

For all the previous experiments we teleported the neural networks at the beginning of the training. Finally, we perform another set of experiments by teleporting the neural network once at epoch $30$. We can clearly see the jump in accuracy right after the teleportation in Fig.~\ref{fig:jump}. And in Fig.~\ref{fig:jump_diffcobs} we compare the training curve of a model with a training were we teleport once at epoch $30$ with different CoB-ranges of $0.7$, $0.8$ and $0.9$. We can see how the bigger the Cob-range the higher the impact on validation accuracy. Experiments of Figs.~\ref{fig:jump} and \ref{fig:jump_diffcobs} where run with a learning rate of $0.0001$ on CIFAR-10.

%We go further in showing that teleportation that is helping the training by producing \textit{pseudo-teleportations}. Given a network $W$ and a random teleportation $V$, we can think of the sphere in weight space with center $W$ and radius vector $\overrightarrow{WV}$. A pseudo-teleportation is another set of weights lying in the boundary of this sphere that is randomly produced with a noise vector. 

%Finally, we produced histograms of an MLP model with ReLU activations before and after teleportation to have a look at the distribution of the weights and to what teleportation does to it, see Fig.~\ref{fig:hist_weights_teleport}.

%Therefore, teleportation allows the exploration of different landscapes that have been proved to be beneficial for training.

%This implies that, even though this training effect of neural teleportation is not addressed in \cite{ArmentaJodoin20}, ...

%\subsection{Speed up of robust training}

%We trained a VGG model with learning rate 0.0001 until it reaches 85\% validation accuracy by teleporting the network at the beginning of the training, and that same model with learning rate 0.01 without teleporting until it reaches the same validation accuracy.

%\subsection{Other activation functions}

%We trained different MLPs with activation functions GELU, tanh, sigmoid, ... using SGD with momentum and a learning rate of 0.001, and compare to what happens if its teleported at the beginning of the training or not.

%-------------------------------------------------------------------------

\section{Conclusion}
In this paper, we provided empirical evidence that neural teleportation can project the weights (and the activation functions) of a network to an infinite amount of places in the weight space while always preserving the network function.  We show that : (1) network teleportation changes the feature maps learned by a network (Fig.~\ref{fig:feature-maps}); (2) It can be used to explore loss level curves; (3) Micro-teleportation vectors are always perpendicular to back-propagated gradients (Fig.~\ref{fig:micro1}); (4) teleportation reduces the landscape flatness (Fig.~\ref{fig:lossLandscape}) and increases the magnitude of the normalized local gradient (Figs.~\ref{fig:grad_change} and \ref{fig:jump_diffcobs}) proportionally to the CoB range $\sigma$ used, and (5) applying \textit{one} teleportation during training accelerates the convergence of gradient descent with and without momentum (Figs.~\ref{fig:training_curves}, \ref{fig:training_curves_otheracts}, \ref{fig:init}, \ref{fig:jump}).   %All these observations are true regardless of the network architecture, the activation functions and the dataset.  As a consequence, we have shown that teleporting a network right before its training accelerates gradient descent.

Although teleportation looks only like a trick to preserve the network function by just rescaling the weights and the activation functions, it is more than a trick as (i) it comes from the foundational definitions and constructions of representation theory, which neural networks exactly satisfy, and (ii) it has unexpected properties on neural network training when applied \textit{once}. The latter underlines the fact that we really don't know the landscape of neural networks and that more tools (or new mathematics) are needed for further understanding. Also, in principle, a neural network and its teleportation should train very similar since they both have the same network function and we have demonstrated that this is not the case.

We conclude that neural teleportation, which is a very simple operation, has remarkable and unexpected properties. We expect these phenomena to motivate the study of the link between neural networks and quiver representations.

Finally, we acknowledge that several different types of change of basis sampling can be performed instead of the uniform distribution around $1$ and $-1$ that we have chosen. In any case, if there are other types of change of basis sampling for which the properties of interest do not hold, then we ask: what makes uniform sampling different?

\bibliography{egbib}

\newpage
\clearpage

\section*{Appendix}

\subsection*{Residual connections and batchnorm}

Here we present an ilustration of how teleportation acts on both residual connections and batchnorm as done by \citep[chap. 5]{ArmentaJodoin20}.

Consider a simple neural network with $5$ neurons where the second and fourth neurons are connected by a residual connection, i.e., its underlying graph (quiver) is of the form
\[
    \begin{tikzpicture}[auto]
       \node (A) at (1,9) {$a$};
       \node (B) at (3,9) {$b$};
       \node (C) at (5,9) {$c$};
       \node (D) at (7,9) {$d$};
       \node (E) at (9,9) {$e$};
       \path[-stealth] 
       %(A) edge [loop above] node {} (A)
       (B) edge [loop above] node {} (B)
       (C) edge [loop above] node {} (C)
       (D) edge [loop above] node {} (D);
       %(E) edge [loop above] node {} (E);
       \draw[->] (A) to  node {$\alpha$} (B);
       \draw[->] (B) to  node {$\beta$} (C);
       \draw[->] (C) to  node {$\gamma$} (D);
       \draw[->] (D) to  node {$\delta$} (E);
       \draw[->] (B) to [ncbar=-1.5em] node[pos=0.5,below] {$\epsilon$} (D); 
    \end{tikzpicture}
\]
and the neural network looks like
\[
    \begin{tikzpicture}[auto]
       \node (A) at (1,9) {$\mathbb{R}$};
       \node (B) at (3,9) {$\mathbb{R}$};
       \node (C) at (5,9) {$\mathbb{R}$};
       \node (D) at (7,9) {$\mathbb{R}$};
       \node (E) at (9,9) {$\mathbb{R}.$};
       \path[-stealth] 
       %(A) edge [loop above] node {$f_a$} (A)
       (B) edge [loop above] node {$f_b$} (B)
       (C) edge [loop above] node {$f_c$} (C)
       (D) edge [loop above] node {$f_d$} (D);
       %(E) edge [loop above] node {$f_e$} (E);
       \draw[->] (A) to  node {$W_\alpha$} (B);
       \draw[->] (B) to  node {$W_\beta$} (C);
       \draw[->] (C) to  node {$W_\gamma$} (D);
       \draw[->] (D) to  node {$W_\delta$} (E);
       \draw[->] (B) to [ncbar=-1.5em] node[pos=0.5,below] {1} (D); 
    \end{tikzpicture}
\]
where $f_b, f_c$ and $f_d$ are activation functions. A CoB of this neural newtork is of the form $\tau = (1, \tau_b, \tau_c, \tau_d, 1)$, so that the teleportation with respect to $\tau$ is the following neural newtork
    \[
    \begin{tikzpicture}[auto]
       \node (A) at (1,9) {$\mathbb{R}$};
       \node (B) at (3,9) {$\mathbb{R}$};
       \node (C) at (5,9) {$\mathbb{R}$};
       \node (D) at (7,9) {$\mathbb{R}$};
       \node (E) at (9,9) {$\mathbb{R}.$};
       \path[-stealth] 
       %(A) edge [loop above] node {$f_a$} (A)
       (B) edge [loop above] node {$\tau_b \cdot f_b$} (B)
       (C) edge [loop above] node {$\tau_c \cdot f_c$} (C)
       (D) edge [loop above] node {$\tau_d \cdot f_d$} (D);
       %(E) edge [loop above] node {$f_e$} (E);
       \draw[->] (A) to  node {$W_\alpha \tau_b$} (B);
       \draw[->] (B) to  node {$W_\beta / \tau_b$} (C);
       \draw[->] (C) to  node {$W_\gamma \tau_d$} (D);
       \draw[->] (D) to  node {$W_\delta / \tau_d$} (E);
       \draw[->] (B) to [ncbar=-1.5em] node[pos=0.5,below] {$\tau_b/\tau_d$} (D); 
    \end{tikzpicture}
    \]
where $\tau_b \cdot f_b(x) = \tau_b f_b(x/\tau_b)$. So if $\tau_b=\tau_d$ then the the teleported network looks as follows
    \[
    \begin{tikzpicture}[auto]
       \node (A) at (1,9) {$\mathbb{R}$};
       \node (B) at (3,9) {$\mathbb{R}$};
       \node (C) at (5,9) {$\mathbb{R}$};
       \node (D) at (7,9) {$\mathbb{R}$};
       \node (E) at (9,9) {$\mathbb{R}.$};
       \path[-stealth] 
       %(A) edge [loop above] node {$f_a$} (A)
       (B) edge [loop above] node {$\tau_b \cdot f_b$} (B)
       (C) edge [loop above] node {$\tau_c \cdot f_c$} (C)
       (D) edge [loop above] node {$\tau_b \cdot f_d$} (D);
       %(E) edge [loop above] node {$f_e$} (E);
       \draw[->] (A) to  node {$W_\alpha \tau_b$} (B);
       \draw[->] (B) to  node {$W_\beta / \tau_b$} (C);
       \draw[->] (C) to  node {$W_\gamma \tau_b$} (D);
       \draw[->] (D) to  node {$W_\delta / \tau_b$} (E);
       \draw[->] (B) to [ncbar=-1.5em] node[pos=0.5,below] {$1$} (D); 
    \end{tikzpicture}
    \]
Therefore, if the CoBs of layers connected with residual connections are equal then the teleportation produced with that CoB will have a residual connection in the same place as the original network.

Consider now the following graph (quiver)
\[
	\begin{tikzpicture}
         	\matrix (m) [matrix of math nodes,row sep=2em,column sep=2em]
          	{
           	  	  & a & & c \\
           	  	 b & & d &  \\
           	};
          	\path[-stealth]
   	     (m-1-2) edge node [below] {} (m-1-4)
       	 (m-2-1) edge node [below] {} (m-1-2)
       	 (m-2-3) edge node [below] {} (m-1-4)
       	 (m-1-2) edge [loop above] node {} (m-1-2)
       	 (m-1-4) edge [loop above] node {} (m-1-4);
	\end{tikzpicture}
\]
We can describe the batchnorm operation on this graph if we consider a layer with only one neuron as follows
\[
	\begin{tikzpicture}
         	\matrix (m) [matrix of math nodes,row sep=2em,column sep=2em]
          	{
           	  	  & \mathbb{R} & & \mathbb{R} \\
           	  	 \mathbb{R} & & \mathbb{R} &  \\
           	};
          	\path[-stealth]
   	     (m-1-2) edge node [below] {$\gamma/\sigma^2$} (m-1-4)
       	 (m-2-1) edge node [below] {$-\mu$} (m-1-2)
       	 (m-2-3) edge node [below] {$\beta$} (m-1-4)
       	 (m-1-2) edge [loop above] node {$1$} (m-1-2)
       	 (m-1-4) edge [loop above] node {$1$} (m-1-4);
	\end{tikzpicture}
\]
where the bottom neurons are considered as bias neurons. It is easy to see that if we feed an input $x$ through this network (i.e., in the top left neuron) we obtain the desired operation $x \mapsto x-\mu \mapsto (x-\mu)(\gamma/\sigma^2)+\beta$. A CoB for this simple layer is of the form $\tau = (\tau_a, \tau_b, \tau_c, \tau_d)$ where $\tau_b=\tau_d=1$ because the bottom neurons are considered as bias neurons. And so a teleportation of the previous batchnorm layer with a CoB $\tau = (\tau_a, 1, \tau_c, 1)$ is of the form
\[
	\begin{tikzpicture}
         	\matrix (m) [matrix of math nodes,row sep=2em,column sep=2em]
          	{
           	  	  & \mathbb{R} & & \mathbb{R} \\
           	  	 \mathbb{R} & & \mathbb{R} &  \\
           	};
          	\path[-stealth]
   	     (m-1-2) edge node [below] {$\gamma \tau_c/\sigma^2 \tau_a$} (m-1-4)
       	 (m-2-1) edge node [below] {$-\mu \tau_a$} (m-1-2)
       	 (m-2-3) edge node [below] {$\beta \tau_c$} (m-1-4)
       	 (m-1-2) edge [loop above] node {$1$} (m-1-2)
       	 (m-1-4) edge [loop above] node {$1$} (m-1-4);
	\end{tikzpicture}
\]
and so if $\tau_a = 1$ then we obtain the layer
\[
	\begin{tikzpicture}
         	\matrix (m) [matrix of math nodes,row sep=2em,column sep=2em]
          	{
           	  	  & \mathbb{R} & & \mathbb{R} \\
           	  	 \mathbb{R} & & \mathbb{R} &  \\
           	};
          	\path[-stealth]
   	     (m-1-2) edge node [below] {$\gamma \tau_c/\sigma^2$} (m-1-4)
       	 (m-2-1) edge node [below] {$-\mu$} (m-1-2)
       	 (m-2-3) edge node [below] {$\beta \tau_c$} (m-1-4)
       	 (m-1-2) edge [loop above] node {$1$} (m-1-2)
       	 (m-1-4) edge [loop above] node {$1$} (m-1-4);
	\end{tikzpicture}
\]
which is a batchnorm layer since the mean and variance $\mu$ and $\sigma^2$ remain the same after teleportation.

\subsection*{Micro-teleportations}

\begin{figure}[tp]
\begin{center}
\includegraphics[scale=0.3]{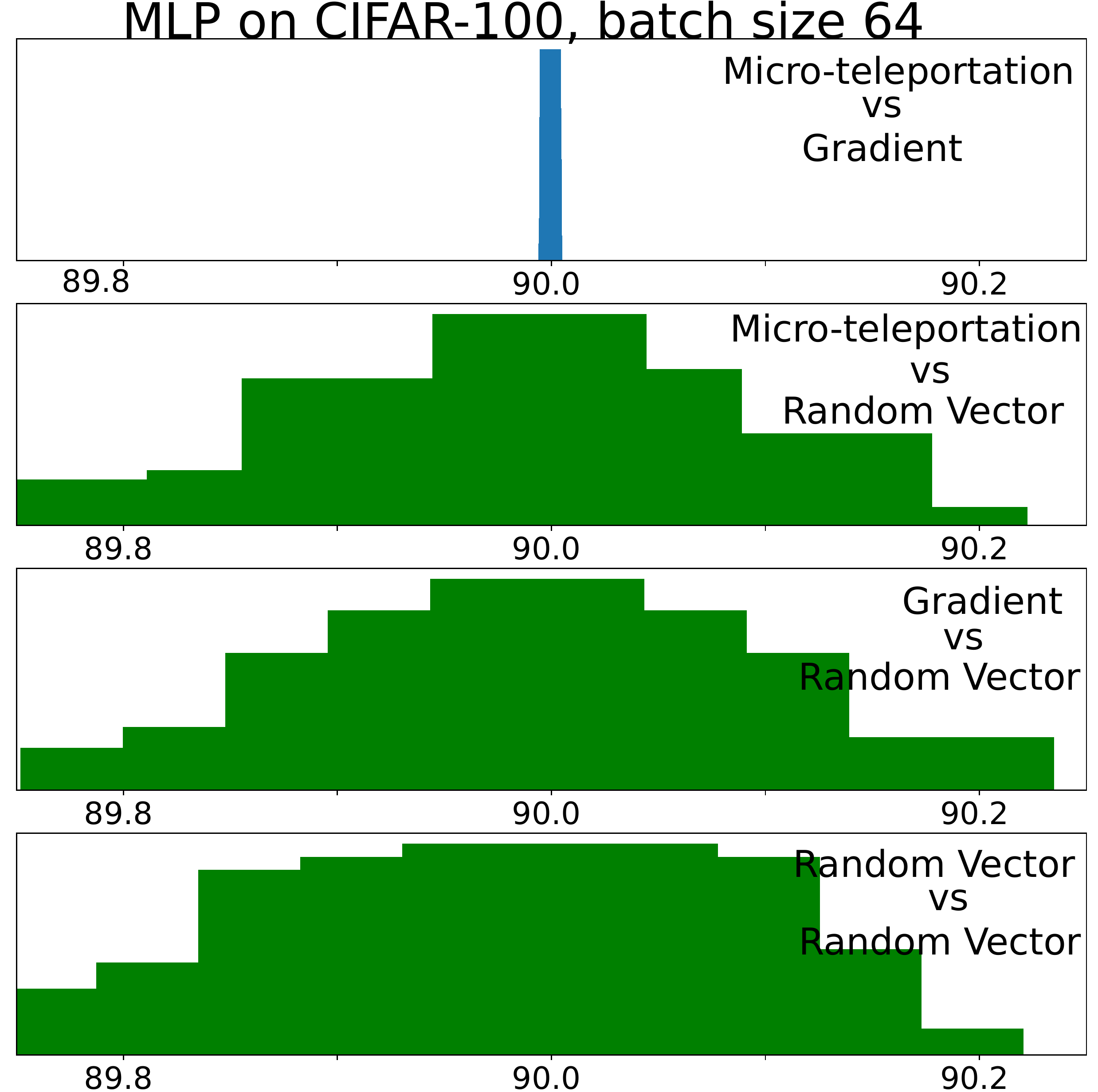}
\includegraphics[scale=0.3]{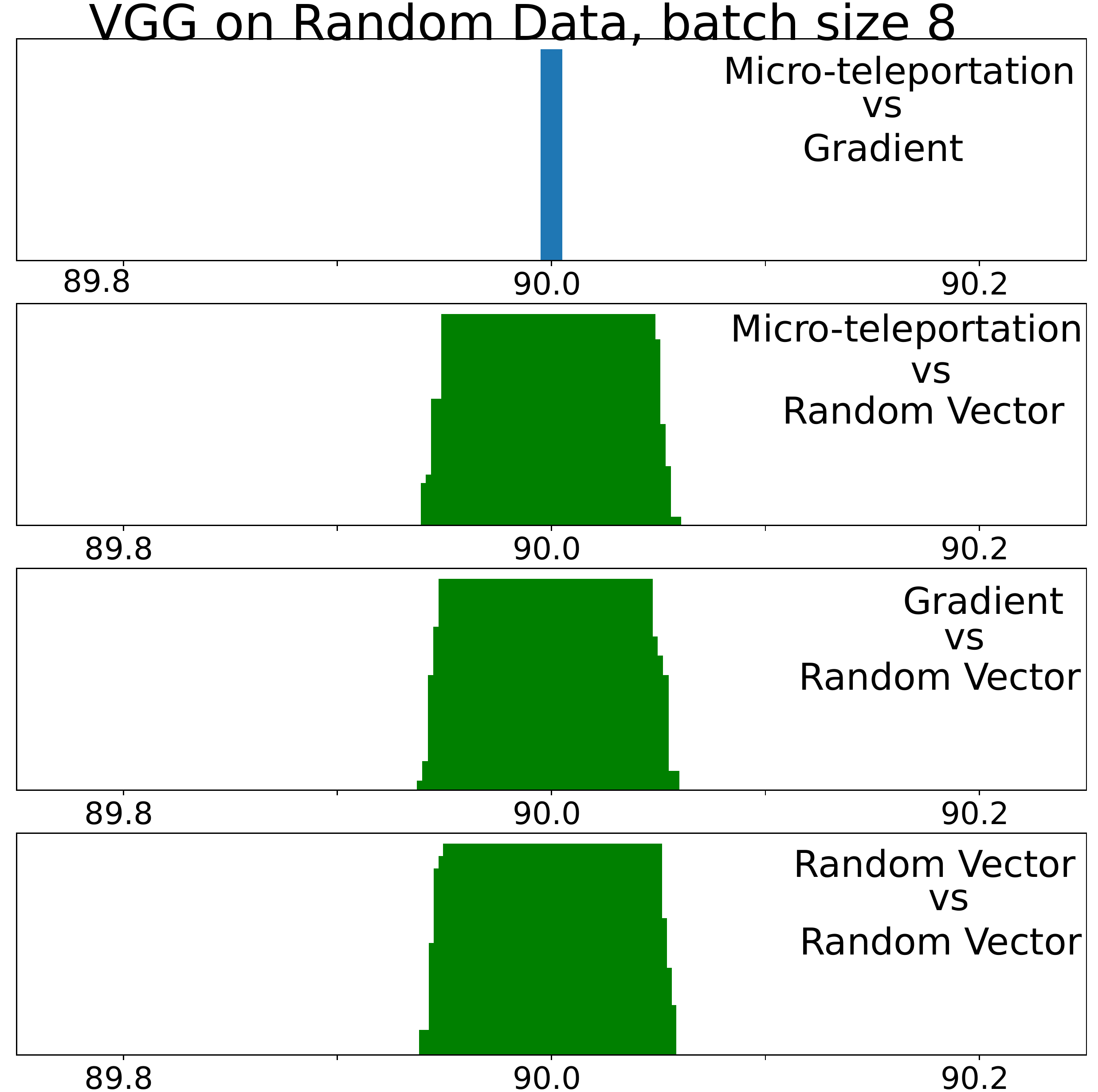}
\end{center}
   \caption{Micro teleportations 
   } \label{fig:micro-1}
\end{figure}

\begin{figure}[tp]
\begin{center}
\includegraphics[width=0.48\linewidth]{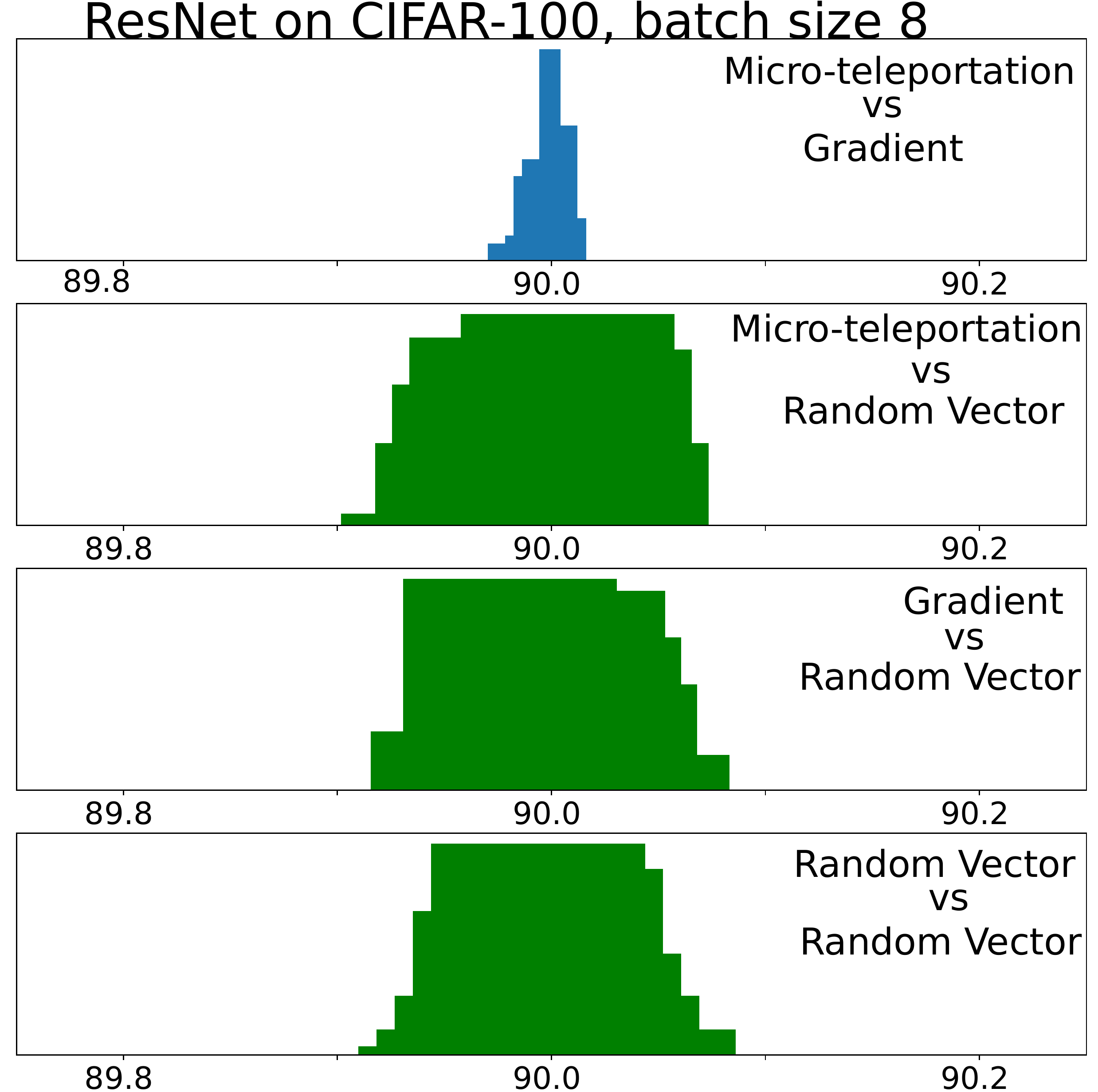}
\includegraphics[width=0.48\linewidth]{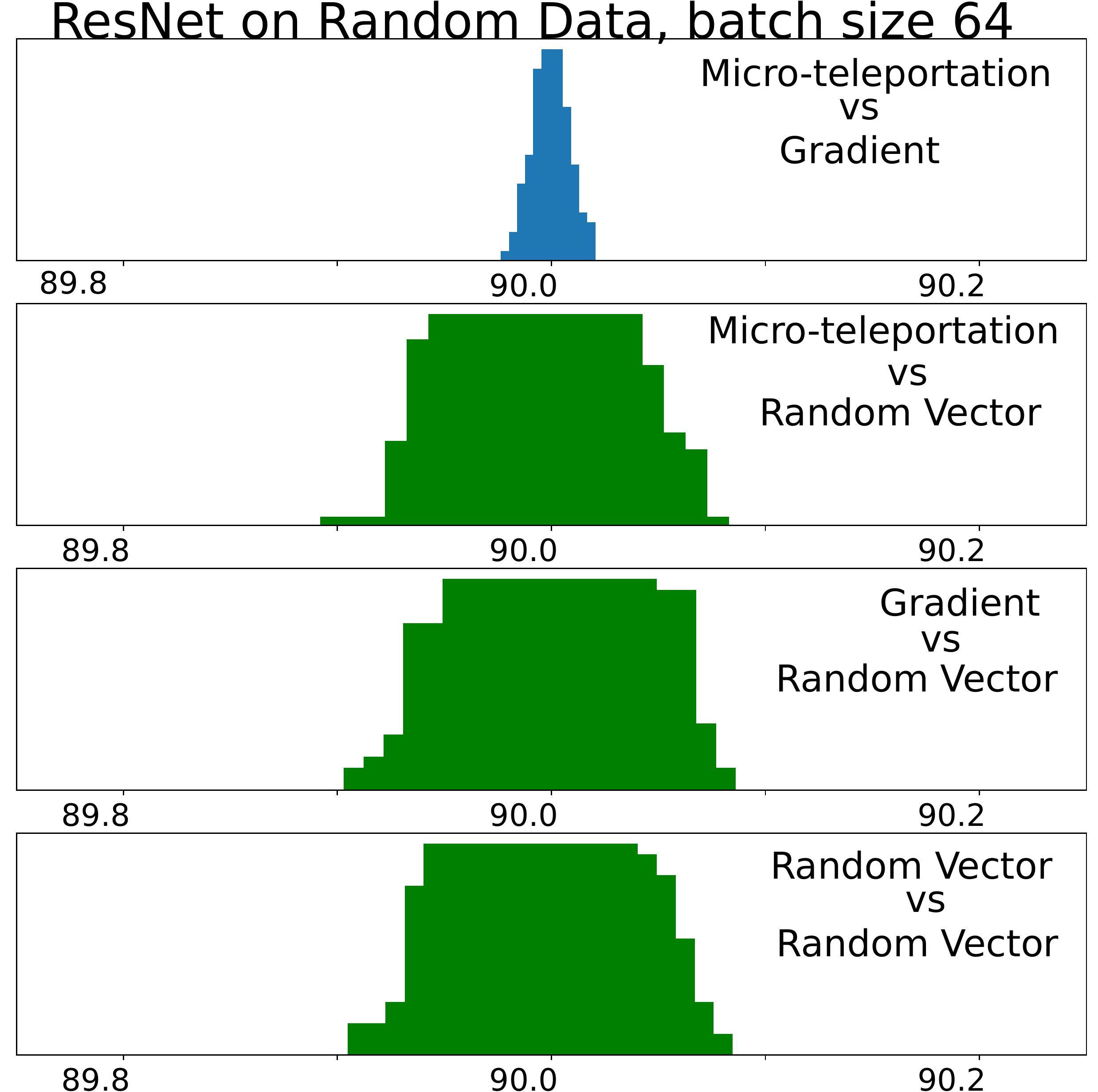}
\end{center}
   \caption{Micro teleportations 
   } \label{fig:micro-2}
\end{figure}

\begin{figure}[tp]
\begin{center}
\includegraphics[scale=0.3]{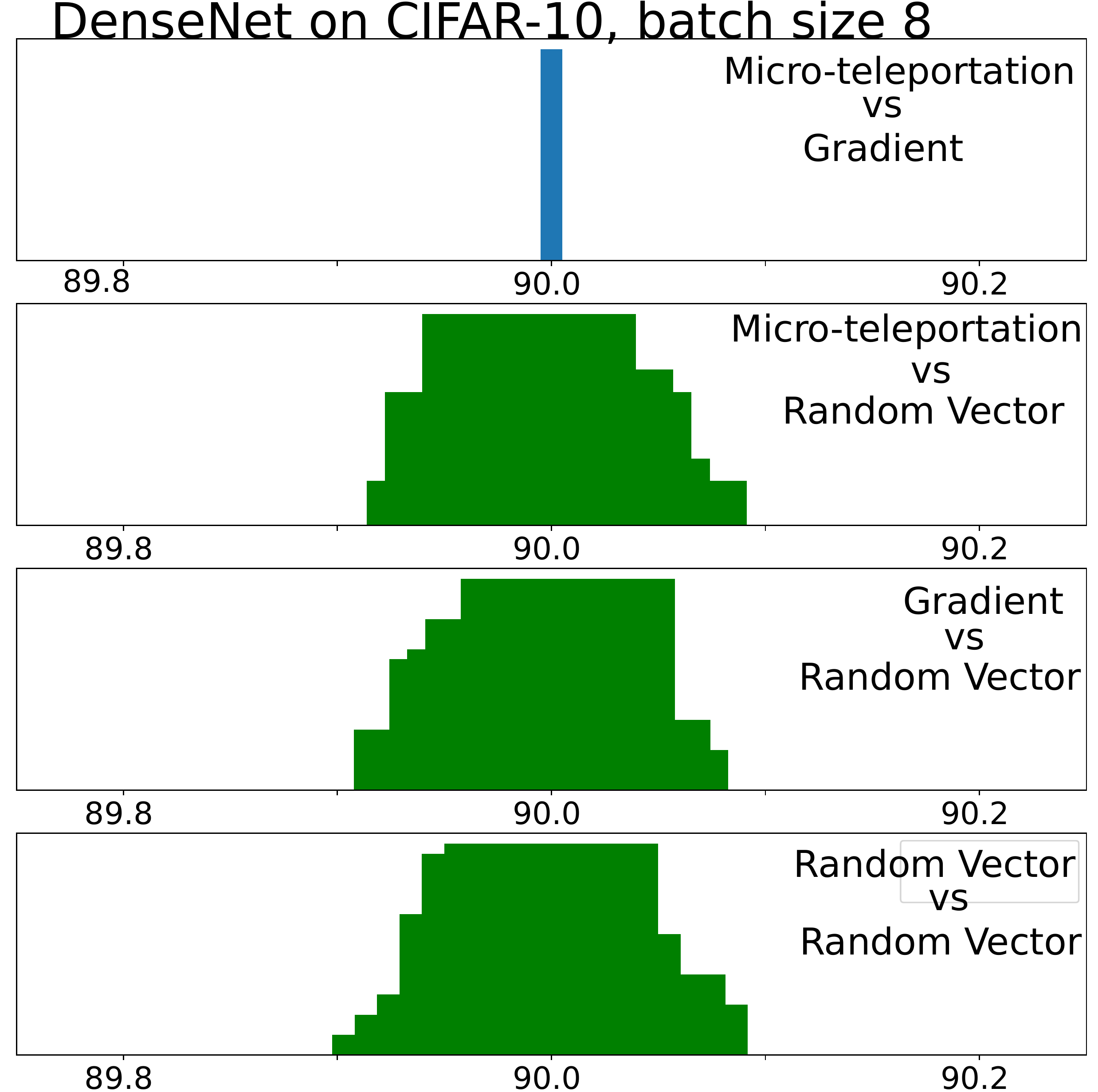}
\includegraphics[scale=0.3]{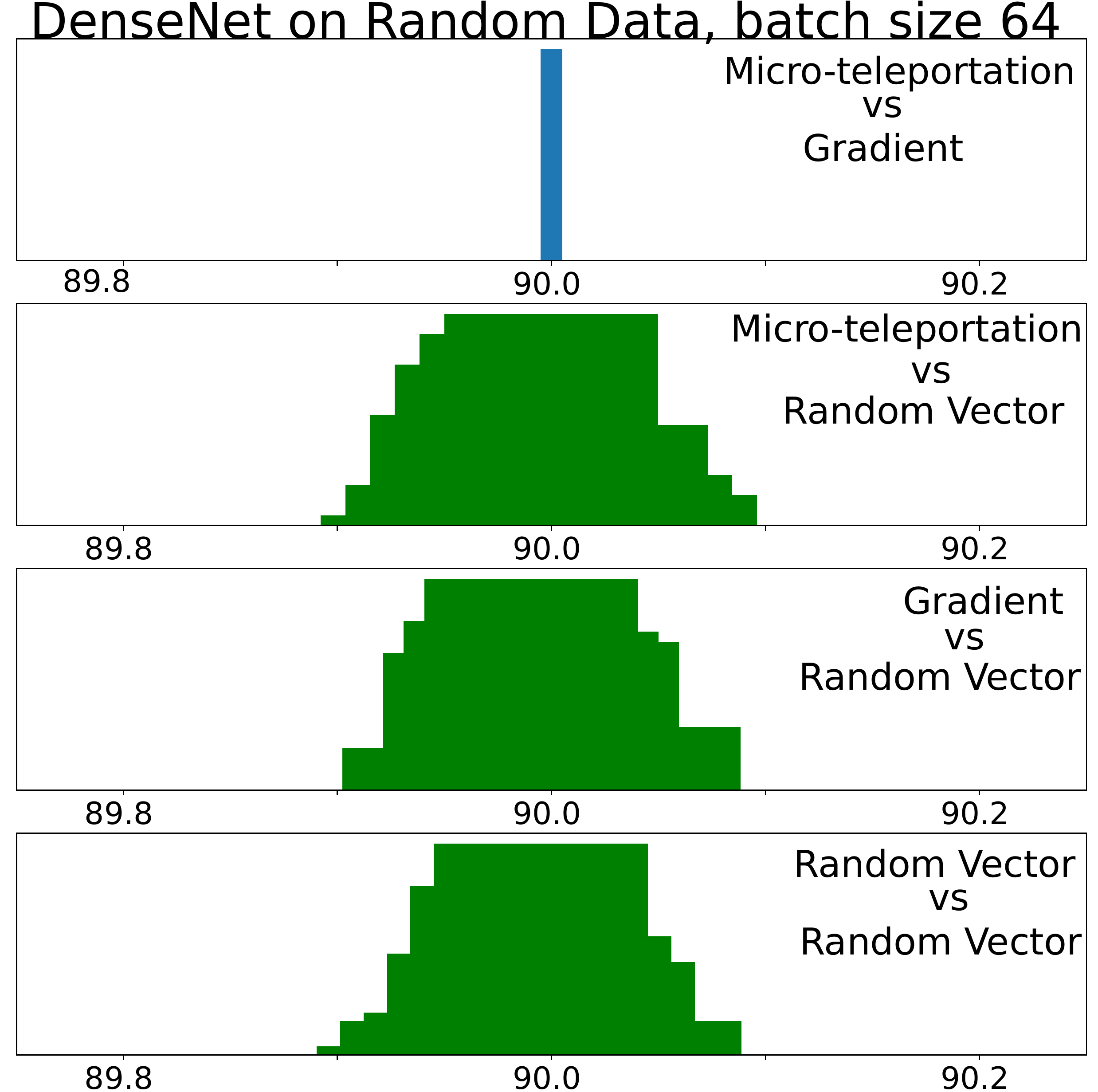}
\end{center}
   \caption{Micro teleportations 
   } \label{fig:micro-3}
\end{figure}

Following section 4.3 on micro-teleportations, we provide more angular histograms between back-propagated gradients and random micro-teleportations of the network. We present here some more histograms for models MLP, VGG, ResNet and DenseNet on datasets CIFAR-10, CIFAR-100 and random data. Each histogram in Figs.~\ref{fig:micro-1}, \ref{fig:micro-2} and \ref{fig:micro-3} was computed with 100 random micro-teleportations with a CoB-range $\sigma=0.001$.

\subsection*{Teleportation and landscape flatness}

\begin{figure*}[tp]
\begin{center}
\includegraphics[width=1\linewidth,height=3.5cm]{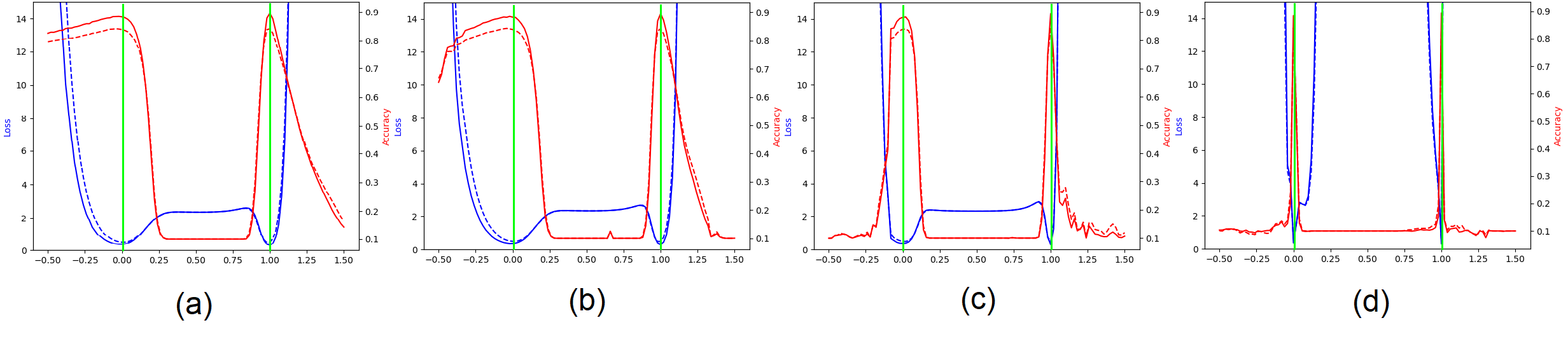}
\end{center}
   \caption{(a) Loss/accuracy profiles obtained by linearly interpolating between two optimized VGG $A$ and $B$. Network $A$ is for $x=0$ and $B$ for $x=1$ (green vertical lines). Dotted lines are for training and solid lines for validation. Remaining plots are similar interpolations but between teleported versions of $A$ and $B$ with CoB range $\sigma$ of (b) $0.6$, (c) $0.9$, and (d) $0.99$.}
\label{fig:interpol2}
\end{figure*}

Following section 4.4, we trained a VGGnet with two different batch sizes and then produced a 1D loss plot by interpolating the two models.  We then re-produced 1D loss plots by teleporting the models.  As shown in Fig.~\ref{fig:interpol2}, neural teleportation has the effect of sharpening the loss landscape.

\subsection*{Back-propagated gradients of a teleportation}

\begin{figure}[tp]
\begin{center}
\includegraphics[width=.24\linewidth]{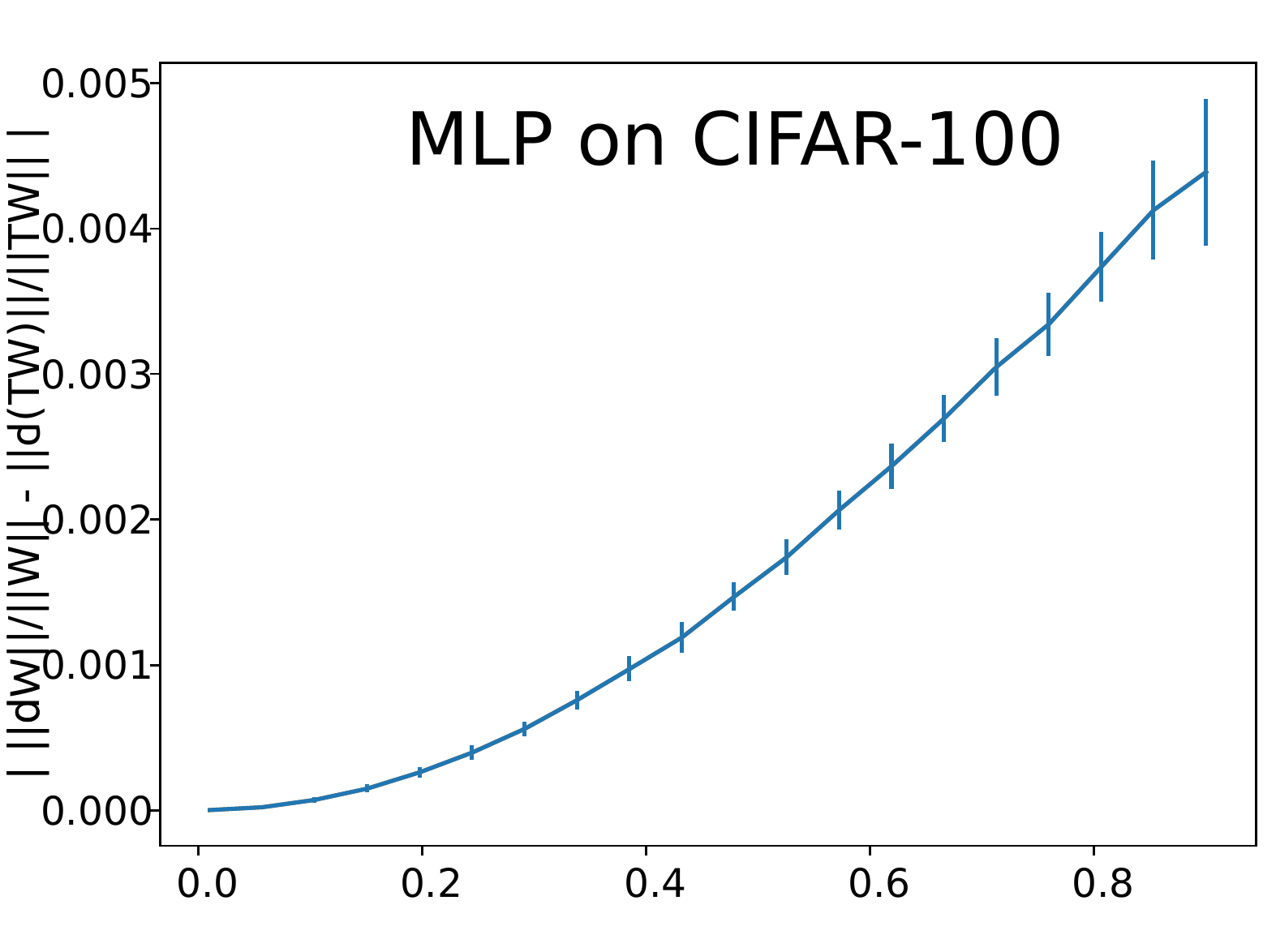}
\includegraphics[width=.24\linewidth]{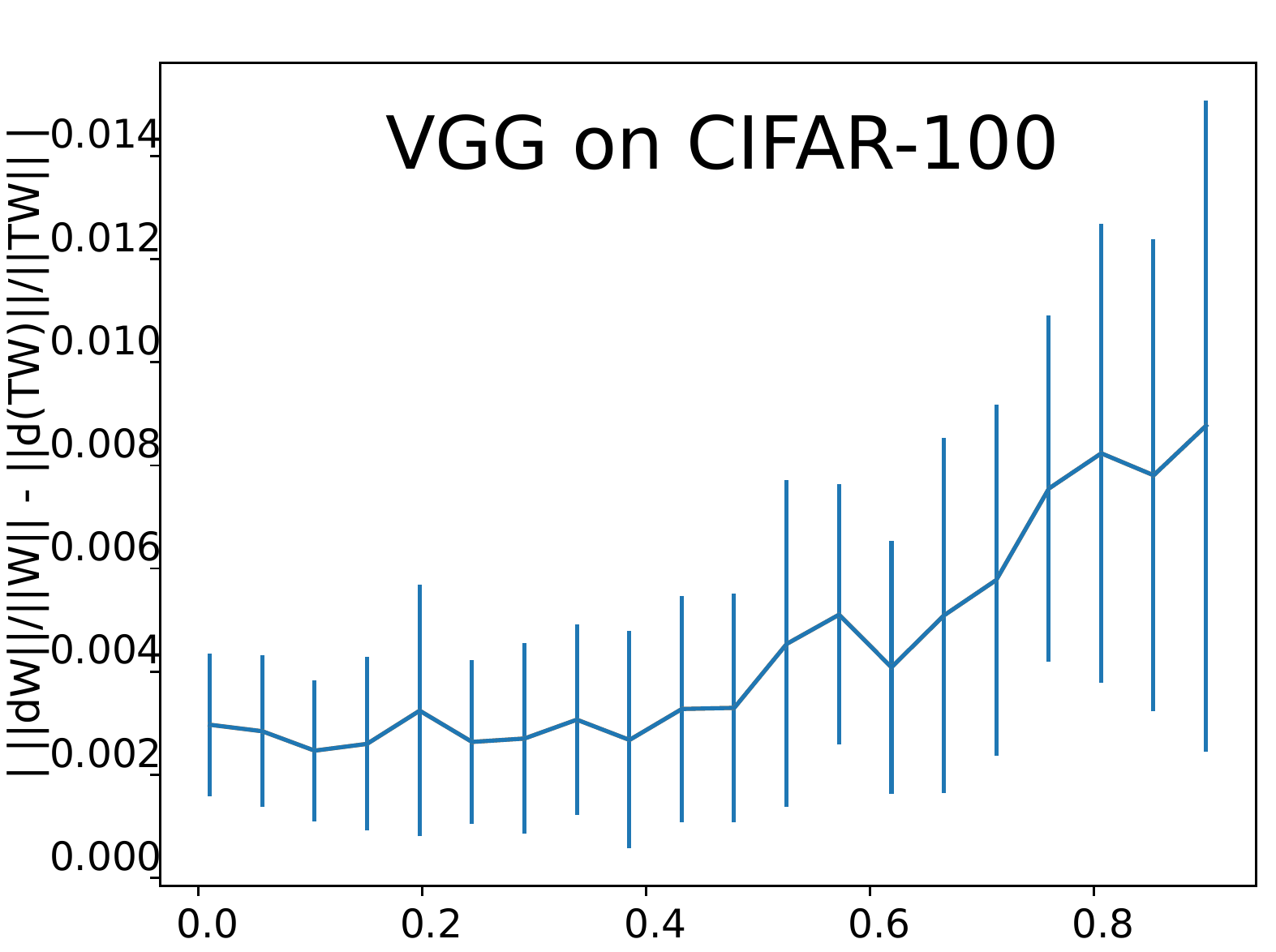}
\includegraphics[width=.24\linewidth]{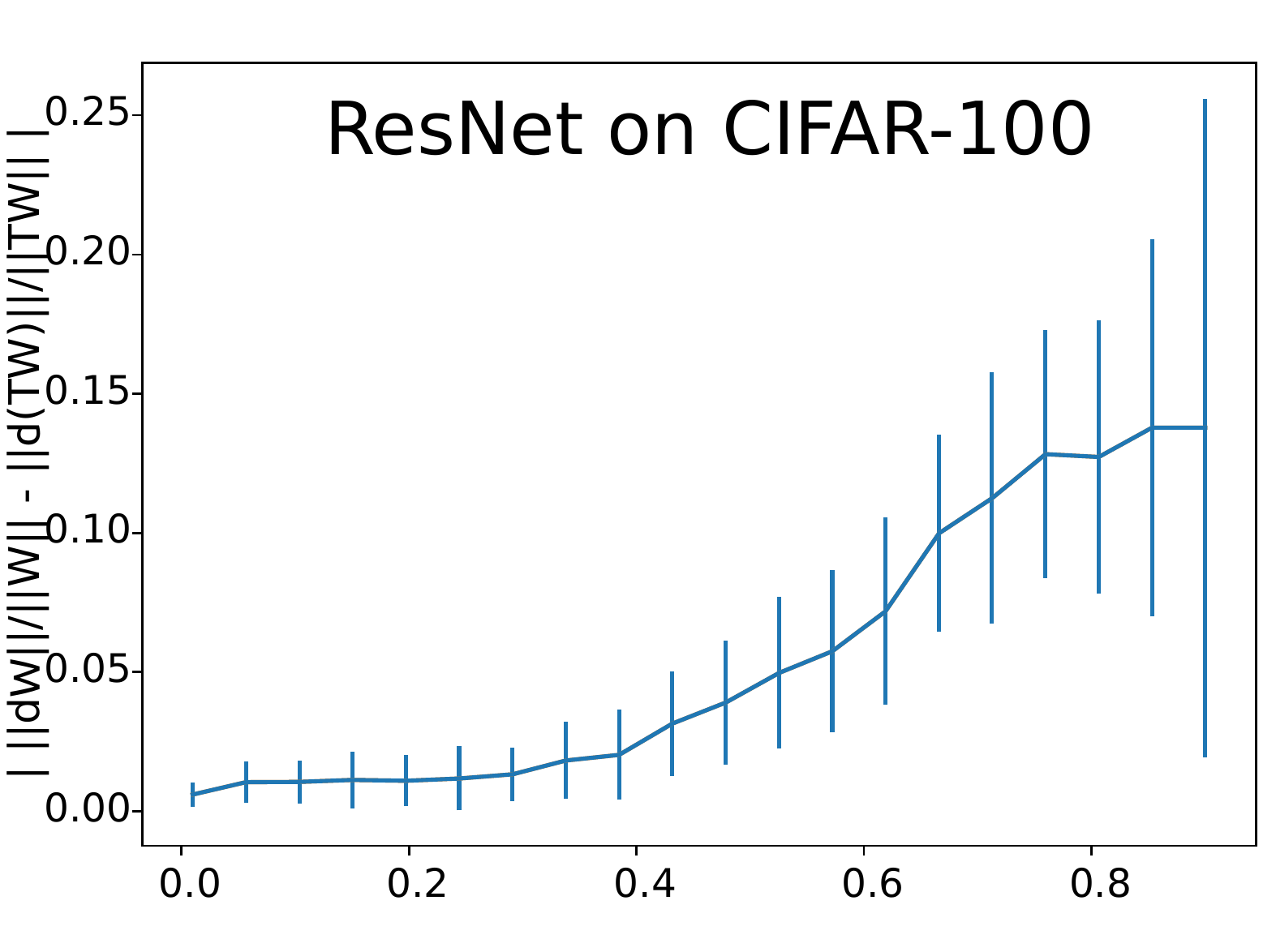}
\includegraphics[width=.24\linewidth]{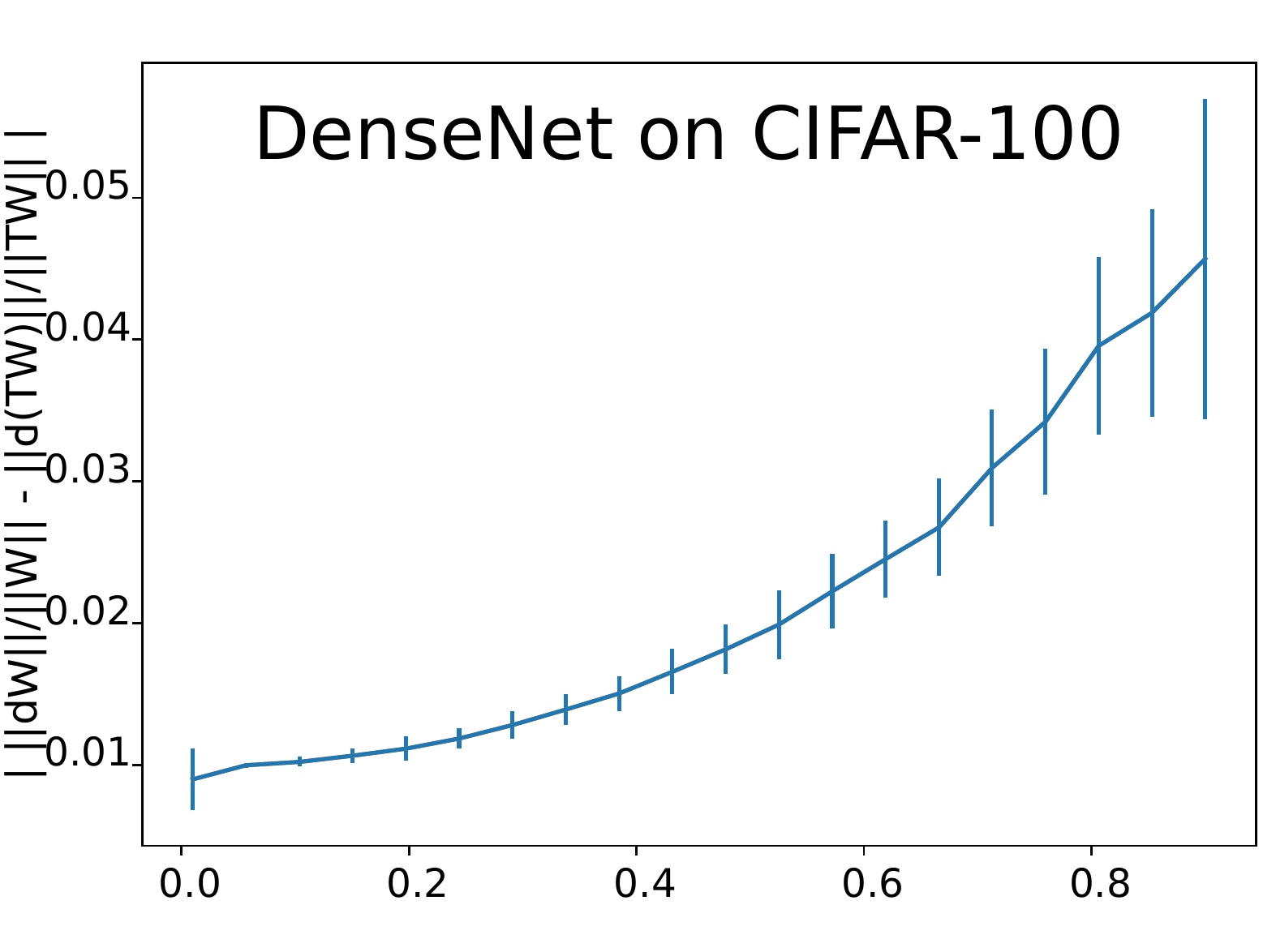}
\end{center}
   \caption{Mean absolute difference ($\pm$ std-dev) between the back-propagated gradients' magnitudes of teleported networks and their original (non-teleported) network.  Larger CoB generate larger gradients. 
   } \label{fig:grad_change2}
\end{figure}

Here we give the proof of Theorem 5.1 and reproduce the experiment shown in Fig.~\ref{fig:grad_change} for the CIFAR-100 dataset. Results are shown in Fig.~\ref{fig:grad_change2}.

%Consider a neural network with weights $W$ and a teleportation $V$. We put in the $x$-axis the CoB-range and we computed the difference
%\[
%    \left| \ \frac{ \|dW\| }{ \|W \|} - \frac{\|  dV  \|}{ \| V \|} \ \right|
%\]
%for $20$ different teleportations for the corresponding CoB-range. Here we show the plots for CIFAR-100.

\subsection*{Proof of Theorem 5.1} 

We will introduce the notation for a forward pass on a neural network $(W,f)$ with $L$ hidden layers without bias vertices for clarity.  For the forward pass of $(W,f)$ we first fix a data sample $(x,t)$ and define the vector of activation outputs of neurons at layer $\ell$ by $a_W^{\left[ \ell \right]}$, and the vector of pre-activations at layer $\ell$ by $z_W^{\left[ \ell \right]}$. We will denote by $f^{[\ell]}$ the vector of activation functions at layer $\ell$. For the input layer we have $a_W^{\left[ 0 \right]} = z_W^{\left[ 0 \right]} = x$. Next, we define inductively
\begin{eqnarray} \label{eq:acts}
    z_W^{\left[ \ell \right]} = W^{\left[ \ell \right]} a_W^{\left[ \ell-1 \right]} \text{   and   } a_W^{\left[ \ell \right]} = f^{\left[ \ell \right]} \left( z_W^{\left[ \ell \right]} \right)
\end{eqnarray}
for every $\ell=1,...,L+1$.

In the case of the backward pass, we denote the vector of derivatives of activations in layer $\ell$ by $df^{\ell}$, and the vector of derivatives with respect to the activation outputs of neurons on layer $\ell$ by $da_W^{\left[ \ell \right]}$. On the output layer we have
\begin{eqnarray} \label{eq:LossDeriv}
    da_W^{\left[ L+1 \right]} = \frac{\partial \Loss }{ \partial a_W^{\left[ L+1 \right]} } (a_W^{\left[ L+1 \right]}, y)    
\end{eqnarray}
where $\Loss$ is the loss function. If we denote by $\odot$ the point-wise (Hadamard) product of vectors of the same size, then inductively from layer $L+1$ down to the input layer we have
\begin{eqnarray} \label{eq:back1}
    dW^{\left[ \ell \right]} = \left( da_W^{\left[ \ell \right]} \odot df^{\left[ \ell \right]} \left( z_W^{\left[ \ell \right]} \right) \right) a_W^{\left[ \ell-1 \right]^T},
\end{eqnarray}
and also that
\begin{eqnarray} \label{eq:back2}
    da_W^{\left[ \ell-1 \right]} = \left(W^{\left[ \ell \right]}\right)^T \left( da_W^{\left[ \ell \right]} \odot df^{\left[ \ell \right]} \left( z_W^{\left[ \ell \right]} \right) \right).
\end{eqnarray}
If $\tau:(W,f) \to (V,g)$ is an isomorphism of neural networks, given by a choice of change of basis in every hidden neuron and we denote by $\tau^{\left[ \ell \right]}$ the vector of change of basis for layer $\ell$, then
\begin{eqnarray} \label{eq:re-scale-weight2}
    V^{\left[ \ell \right]} = \frac{1}{\tau^{\left[ \ell -1 \right]}} \bullet W^{\left[ \ell \right]} \bullet \tau^{\left[ \ell \right]}.
\end{eqnarray}
and for the activation functions we have
\begin{eqnarray} \label{eq:tel-act-vect}
    g^{\left[ \ell \right]}(x) = f^{\left[ \ell \right]}\left( x \bullet \frac{1}{\tau^{\left[ \ell \right]}} \right) \bullet \tau^{\left[ \ell \right]},
\end{eqnarray}
Where the operation $\frac{1}{\tau^{\left[ \ell -1 \right]}} \bullet -$ on the left of a matrix, multiplies its columns by the coordinate values of vector $\frac{1}{\tau^{\left[ \ell -1 \right]}}$. While the operation $- \bullet \tau^{\left[ \ell \right]}$ on the right of a matrix, multiplies its rows by the coordinate values of the vector $ \tau^{\left[ \ell \right]}$.
By transposing Eq.~\ref{eq:re-scale-weight2}, we obtain
\begin{eqnarray} \label{eq:tel-weights-transp}
    \left(  V^{\left[ \ell \right]} \right)^T = \tau^{\left[ \ell \right]} \bullet \left(W^{\left[ \ell \right]}\right)^T \bullet \frac{1}{\tau^{\left[ \ell -1 \right]}}.
\end{eqnarray}

By the chain rule and Eq.~\ref{eq:tel-act-vect} we can see that
\begin{eqnarray} \label{eq:tel-der-act}
    dg^{\left[ \ell \right]}(x) = df^{\left[ \ell \right]} \left( x \bullet \frac{1}{\tau^{\left[ \ell \right]}} \right).
\end{eqnarray}
Also, from the proof of Theorem 4.9 \cite{ArmentaJodoin20},
\begin{eqnarray} \label{eq:tel-preact}
    z_V^{\left[ \ell \right]}=z_W^{\left[ \ell \right]} \bullet \tau^{\left[ \ell \right]} \ \ \ \text{   and   } \ \ \ a_V^{\left[ \ell \right]}=a_W^{\left[ \ell \right]} \bullet \tau^{\left[ \ell \right]}.
\end{eqnarray}

\begin{figure*}[tp]
\begin{center}
\includegraphics[width=0.40\linewidth]{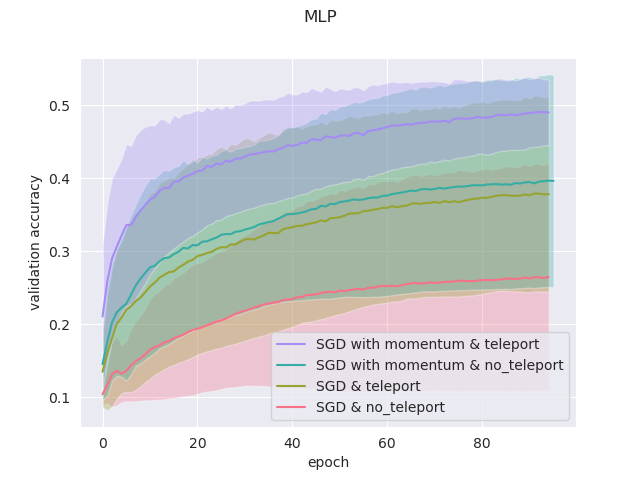}
\includegraphics[width=0.40\linewidth]{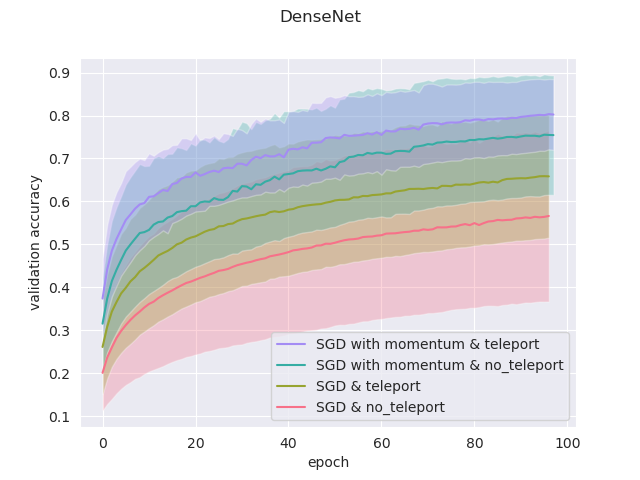}

\includegraphics[width=0.40\linewidth]{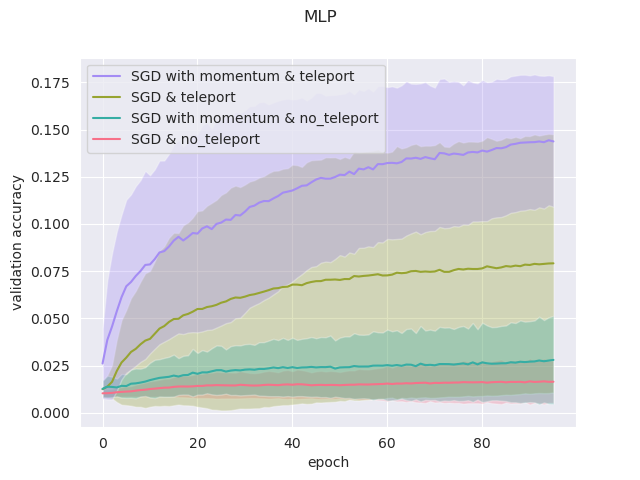}
\includegraphics[width=0.40\linewidth]{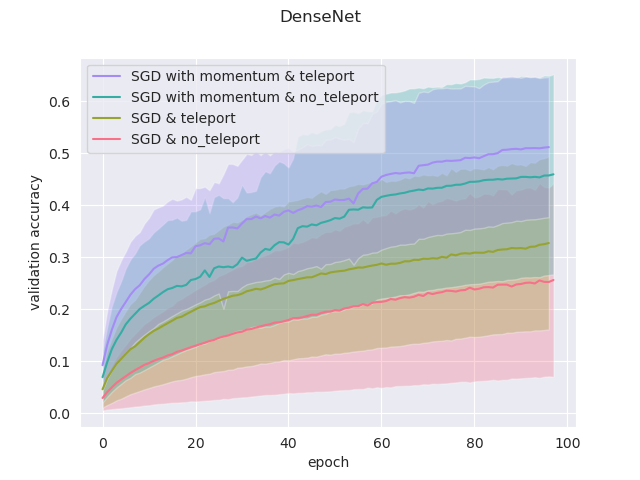}

\end{center}
   \caption{Validation accuracies for MLP and DenseNet on CIFAR-10 [top row] and CIFAR-100 [bottom row]. The curves are produced by averaging over 5 runs for all learning rates ($0.01$, $0.001$ and $0.0001$) with shades representing $\pm \ std-dev$.\vspace{-0.75cm} }
\label{fig:train_apx_1}
\end{figure*}

\begin{figure*}[tp]
    \begin{center}
    \includegraphics[width=0.40\linewidth]{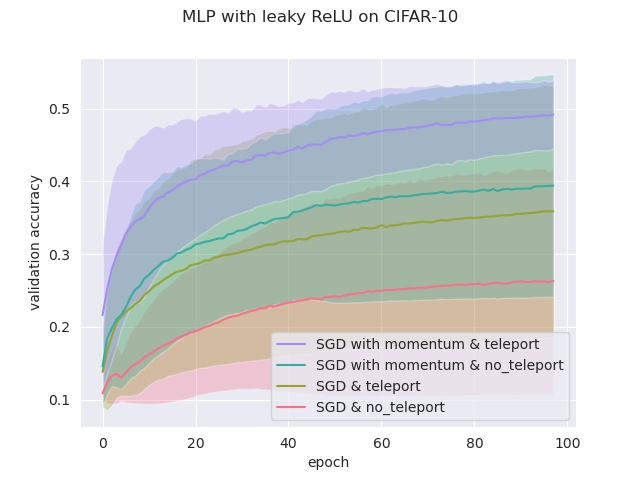}
    \includegraphics[width=0.40\linewidth]{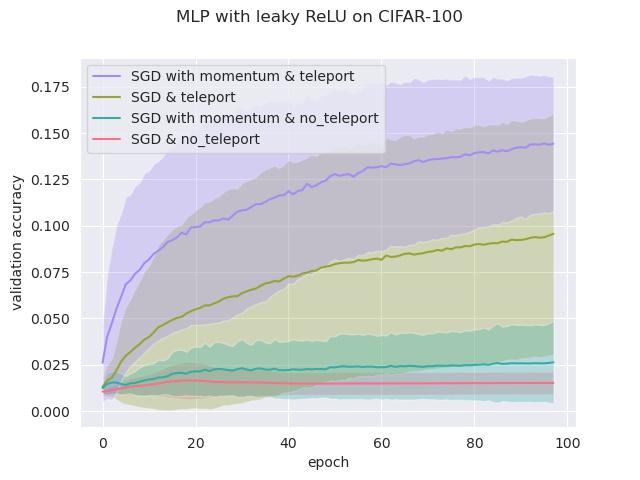}
    
    \includegraphics[width=0.40\linewidth]{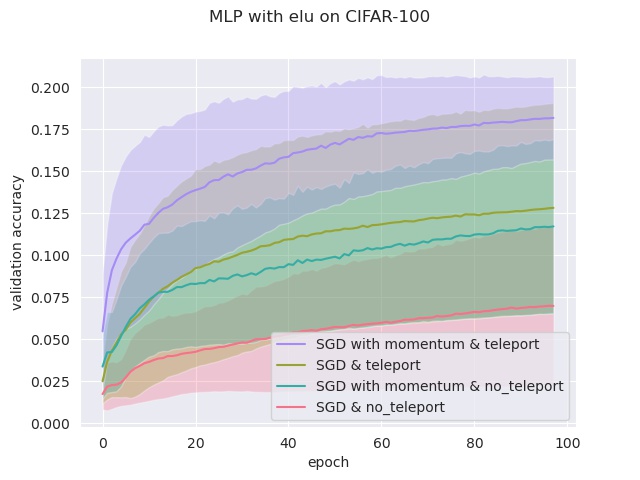}
    \includegraphics[width=0.40\linewidth]{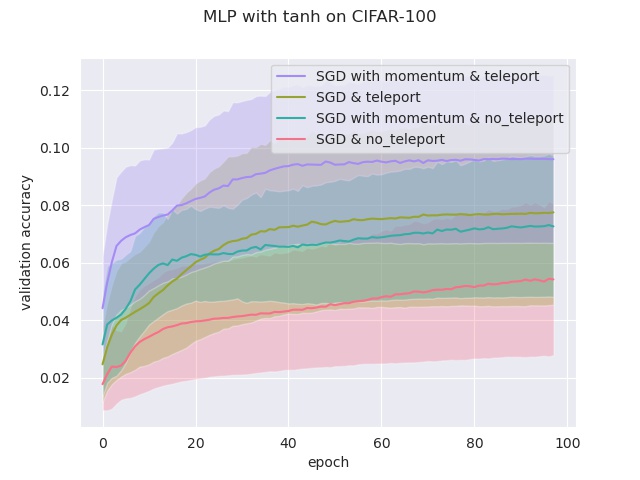}
    
    \end{center}
       \caption{Validation plots produced over 5 runs over three learning rates ($0.01$, $0.001$ and $0.0001$) of MLPs. The top left plot is on CIFAR-10 and the rest on CIFAR-100. }
    \label{fig:train_apx_2}
\end{figure*}

\begin{figure*}[tp]
\begin{center}
\includegraphics[width=0.40\linewidth]{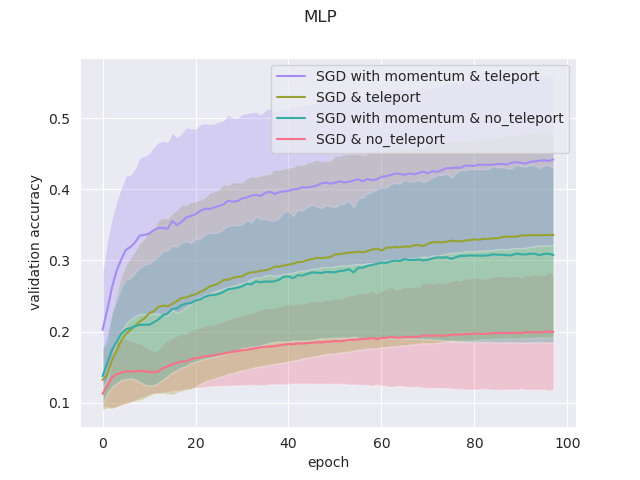}
\includegraphics[width=0.40\linewidth]{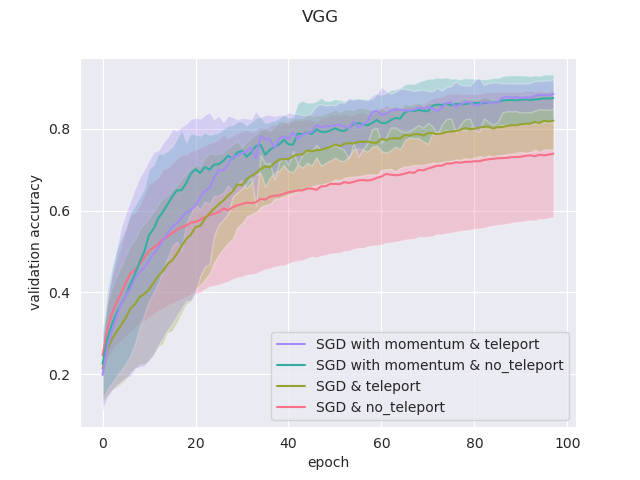}

\includegraphics[width=0.40\linewidth]{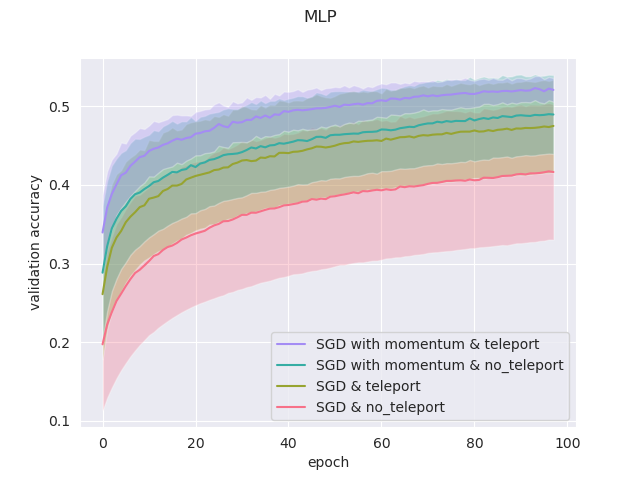}
\includegraphics[width=0.40\linewidth]{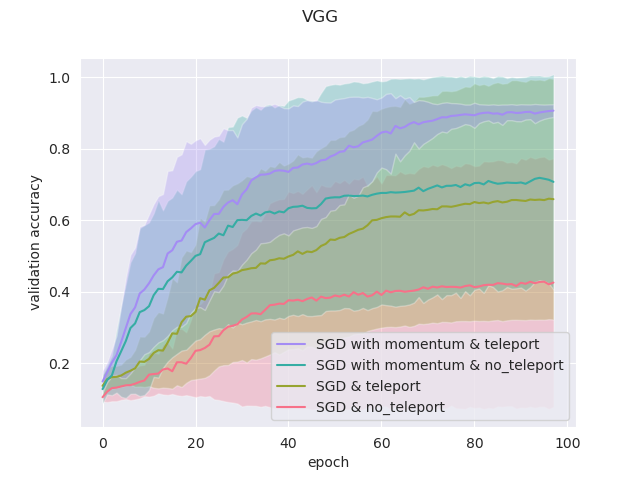}

\end{center}
   \caption{Validation plots produced over $5$ runs over three learning rates ($0.01$, $0.001$ and $0.0001$) of the four models MLP and VGG with normal [top row] and Xavier [bottom row] initializations on CIFAR-10.}
\label{fig:train_apx_3}
\end{figure*}

\begin{figure*}[tp]
\begin{center}

\includegraphics[width=0.40\linewidth]{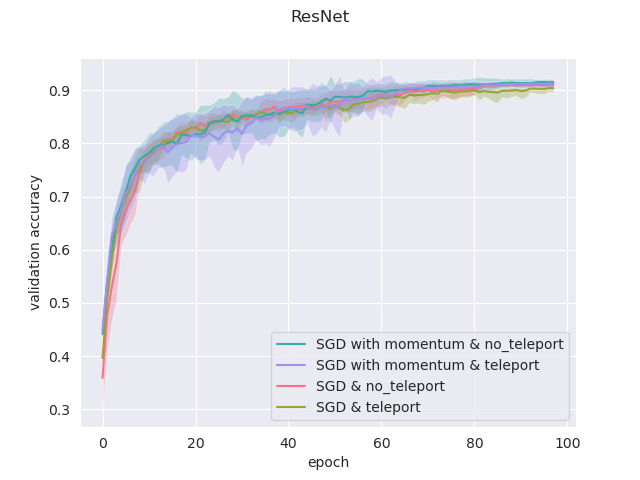}
\includegraphics[width=0.40\linewidth]{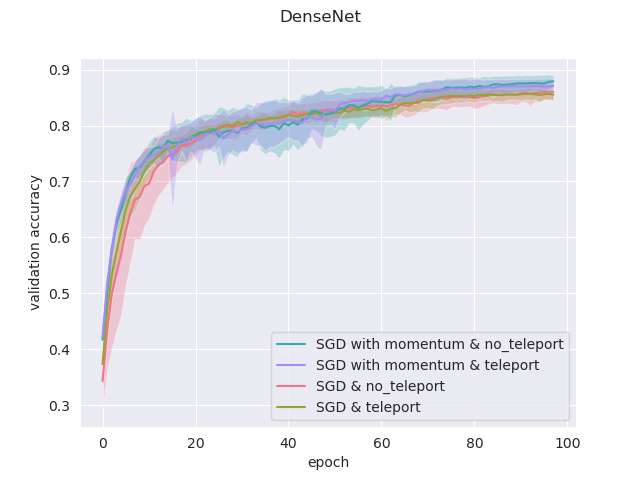}

\end{center}
   \caption{Validation plots produced over $5$ runs over three learning rates ($0.01$, $0.001$ and $0.0001$) of the four models MLP and VGG with normal [top row] and Xavier [bottom row] initializations on CIFAR-10.}
\label{fig:train_apx_4}
\end{figure*}

\begin{figure*}[tp]
    \begin{center}
    \includegraphics[width=0.40\linewidth]{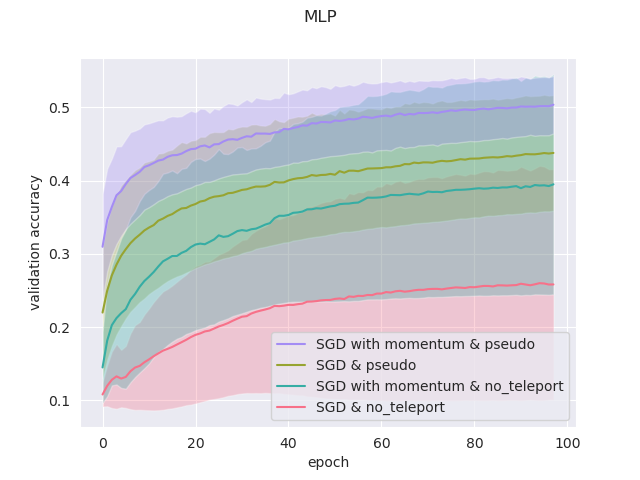}
    \includegraphics[width=0.40\linewidth]{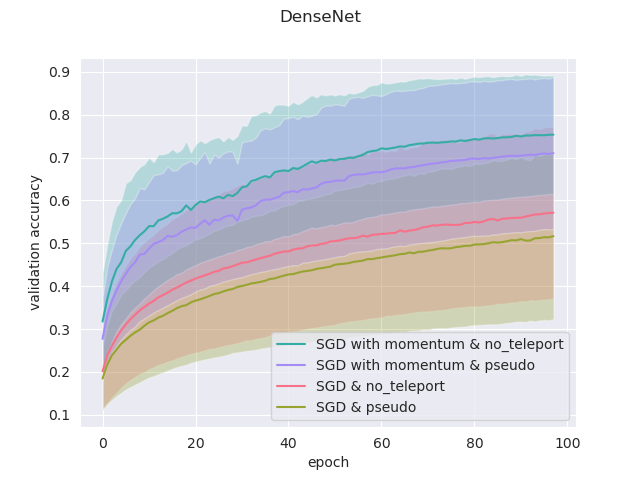}
    
    \includegraphics[width=0.40\linewidth]{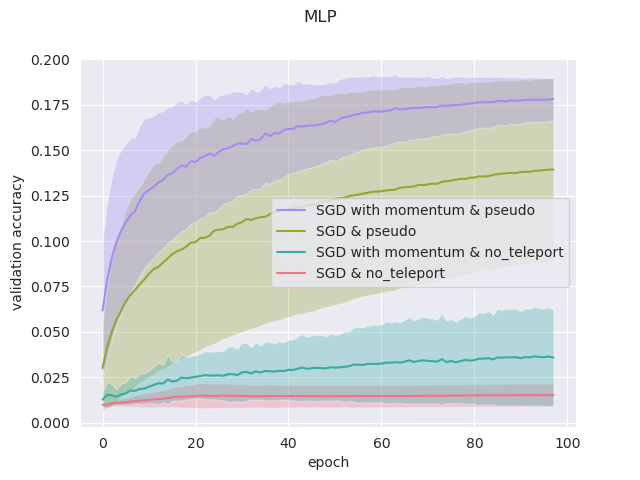}
    \includegraphics[width=0.40\linewidth]{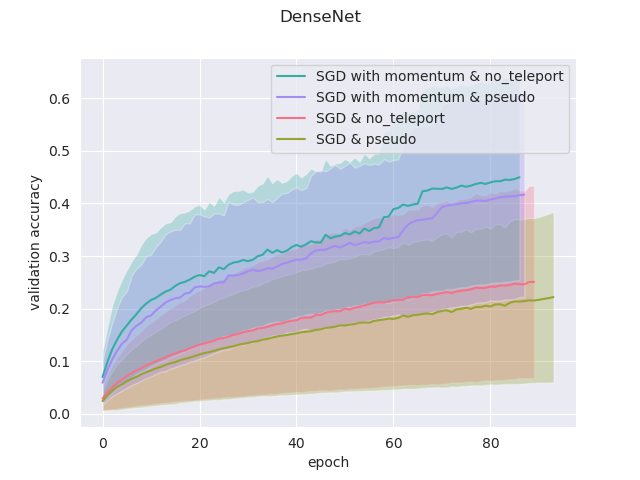}
    
    \end{center}
       \caption{Validation plots produced over $5$ runs over three learning rates ($0.01$, $0.001$ and $0.0001$) of the models MLP and DenseNet with pseudo-teleportation on CIFAR-10 [top row] and CIFAR-100 [bottom row]. }
       \label{fig:train_apx_5}
\end{figure*}

\begin{figure*}[tp]
    \begin{center}
    \includegraphics[width=0.40\linewidth]{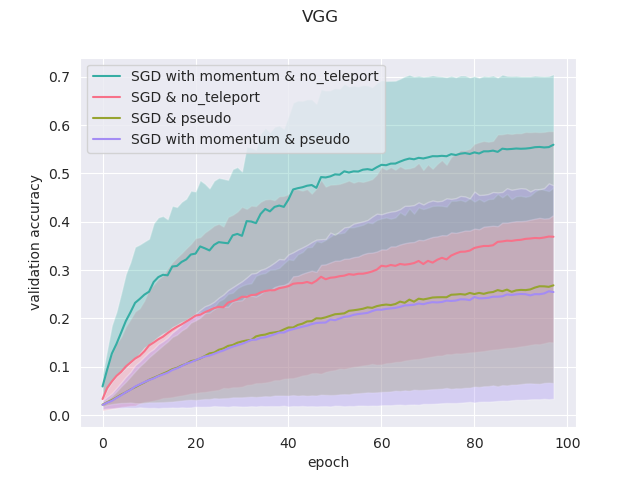}
    \includegraphics[width=0.40\linewidth]{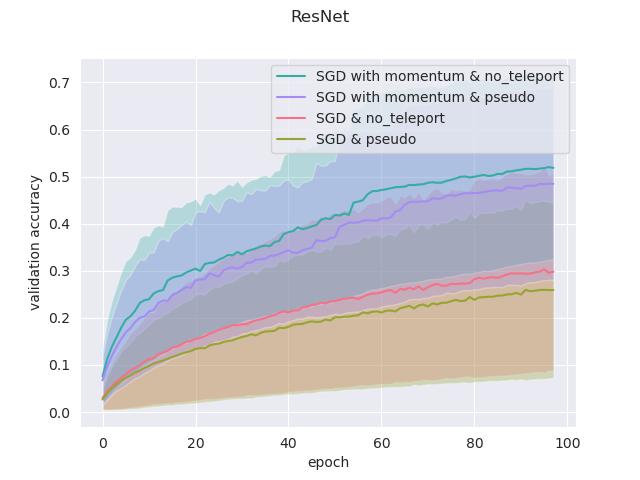}

    \end{center}
       \caption{Validation plots produced over $5$ runs over three learning rates ($0.01$, $0.001$ and $0.0001$) of the models VGG and ResNet with pseudo-teleportation on CIFAR-100. }
       \label{fig:train_apx_6}
\end{figure*}

\begin{Theorem} Let $(V,g)$ be a teleportation of the neural network $(W,f)$ with respect to the CoB $\tau$. Then 
    \[
        dV^{\left[ \ell \right]} = \tau^{\left[ \ell - 1 \right]} \bullet  dW^{\left[ \ell \right]} \bullet \frac{1}{\tau^{\left[ \ell \right]}} 
   \]
    for every layer $\ell$ of the network. 
\end{Theorem}

\textbf{Proof:} We proceed by induction on the steps of the backward pass of the network $(V,g)$. By Theorem 4.9 of \cite{ArmentaJodoin20}, both networks $(W,f)$ and $(V,g)$ give the same output $a_W^{[L+1]} = a_V^{[L+1]}$, so by Eq.~\ref{eq:LossDeriv} we get that $da_W^{\left[ L+1 \right]}=da_V^{\left[ L+1 \right]}$. We can substitute this together with Eq.~\ref{eq:tel-preact} into Eq.~\ref{eq:back1} applied to the neural network $(V,g)$ to obtain
\[
    \begin{array}{lcl}
         dV^{\left[ L+1 \right]} & = & da_V^{\left[ L+1 \right]} a_V^{\left[ L \right]^T} \\
         & = &  da_W^{\left[ L+1 \right]} \left( a_W^{\left[ L \right]} \bullet \tau^{\left[ L \right]} \right)^T \\
         & = & dW^{\left[ L+1 \right]} \bullet \frac{1}{\tau^{\left[ L \right]}}. 
    \end{array}
\]
Now, the derivative of the loss function with respect to the activation outputs of layer $L$ in $(V,g)$ can be written analogously to Eq.~\ref{eq:back2}, in which we substitute Eq.~\ref{eq:tel-weights-transp} taking into account that the CoB for the output layer is given by $1$'s, so we get 
\[ 
    \begin{array}{lcl}
         da_V^{\left[ L \right]} & = & \left( V^{\left[ L+1 \right]} \right)^T da_V^{\left[ L+1 \right]} \\
         & = &  \left( W^{\left[ L+1 \right]} \right)^T \bullet \frac{1}{\tau^{\left[ L \right]}} da_W^{\left[ L+1 \right]} \\
         & = & da_W^{\left[ L \right]} \bullet \frac{1}{\tau^{\left[ L \right]}}.
    \end{array}  %\vspace{-0.2cm}
\]
At layer $L$ we get that 
\[
    \begin{array}{l}
        dV^{[L]}  =  \left( da_V^{\left[ L \right]} \odot dg^{\left[ L \right]}\left( z_V^{\left[ L \right]} \right) \right) a_V^{\left[ L-1 \right]^T}  \\ 
        =  \left( da_W^{\left[ L \right]} \bullet \frac{1}{\tau^{\left[ L \right]}} \right) \odot df^{\left[ L \right]} \left( z_W^{[L]} \bullet \tau^{[L]} \bullet \frac{1}{\tau^{\left[ L \right]}} \right) a_V^{\left[ L-1 \right]^T}  \\ 
        = \left( da_W^{[L]} \odot df^{[L]}\left( z_W^{[L]} \right) \bullet \frac{1}{\tau^{[L]}} \right) \left( a_W^{[L-1]} \bullet \tau^{[L-1]} \right)^T \\ 
        = \left( da_W^{[L]} \odot df^{[L]}\left( z_W^{[L]} \right) \bullet \frac{1}{\tau^{[L]}} \right) \left( \tau^{[L-1]} \bullet a_W^{[L-1]^T} \right) \\ 
        = \tau^{[L-1]} \bullet \left( da_W^{[L]} \odot df^{[L]}\left( z_W^{[L]} \right) a_W^{[L-1]^T} \right) \bullet \frac{1}{\tau^{[L]}} \\ 
        = \tau^{[L-1]} \bullet dW^{[L]} \bullet \frac{1}{\tau^{[L]}}.
    \end{array}
\]
We then observe that
\[
    \begin{array}{l}
        da_V^{[L-1]}  =  \left( V^{[L]} \right)^T  da_V^{[L]}  \\ 
          =  \left( \tau^{\left[ L \right]} \bullet \left(W^{\left[ L \right]}\right)^T \bullet \frac{1}{\tau^{\left[ L -1 \right]}} \right) \left( da_W^{[L]} \bullet \frac{1}{\tau^{[L]}} \right) \\ 
          =  \left(W^{\left[ L \right]}\right)^T da_W^{[L]} \bullet \frac{1}{\tau^{[L-1]}} \\ 
          = da_W^{[L-1]} \bullet \frac{1}{\tau^{[L-1]}}.
    \end{array}
\]
For the inductive step, we assume the result to be true for layers $L+1,...,\ell+1$. We will prove the result holds for layer $\ell$. Indeed,
\[
    \begin{array}{l}
         dV^{\left[ \ell \right]}  =  \left( da_V^{\left[ \ell \right]} \odot dg^{\left[ \ell \right]}\left( z_V^{\left[ \ell \right]} \right) \right) a_V^{\left[ \ell-1 \right]^T} \\ \\
          =  \left( da_W^{\left[ \ell \right]} \bullet \frac{1}{\tau^{\left[ \ell \right]}} \right) \odot df^{\left[ \ell \right]}\left( z_W^{\left[ \ell \right]} \bullet \tau^{\left[ \ell \right]} \bullet \frac{1}{\tau^{\left[ \ell \right]}} \right) a_V^{\left[ \ell-1 \right]^T} \\ \\
          =  \left( da_W^{\left[ \ell \right]} \odot df^{\left[ \ell \right]}\left( z_W^{\left[ \ell \right]} \right) \bullet \frac{1}{\tau^{\left[ \ell \right]}} \right) \left( a_W^{[\ell-1]} \bullet \tau^{[\ell-1]} \right)^T    \\ \\
          =  \left( da_W^{\left[ \ell \right]} \odot df^{\left[ \ell \right]}\left( z_W^{\left[ \ell \right]} \right)  \right) \left(\tau^{\left[ \ell-1 \right]} \bullet a_W^{\left[ \ell-1 \right]^T} \right) \bullet \frac{1}{\tau^{\left[ \ell \right]}} \\ \\
          =  \tau^{\left[ \ell-1 \right]} \bullet \left( da_W^{\left[ \ell \right]} \odot df^{\left[ \ell \right]}\left( z_W^{\left[ \ell \right]} \right)  \right)  a_W^{\left[ \ell-1 \right]^T} \bullet \frac{1}{\tau^{\left[ \ell \right]}} \\ \\
          =  \tau^{\left[ \ell-1 \right]} \bullet dW^{\left[ \ell \right]} \bullet \frac{1}{\tau^{\left[ \ell \right]}}.
    \end{array}
\]
It can be appreciated that back-propagation on $(V,g)$ computes a re-scaling of the gradient of $(W,f)$ by the CoB, just as claimed in the statement of the theorem$._\blacksquare$

\subsection*{Gradient descent speed up}

Following section~\ref{sec:optimization}, we present Figs.~\ref{fig:train_apx_1}, \ref{fig:train_apx_2}, \ref{fig:train_apx_3}, \ref{fig:train_apx_4}, \ref{fig:train_apx_5} with the training curves for the remaining models and datasets of our training experiments.

%{\noindent \em Remainder omitted in this sample. See http://www.jmlr.org/papers/ for full paper.}

% Acknowledgements should go at the end, before appendices and references

%\acks{We would like to acknowledge support for this project
%5from the National Science Foundation (NSF grant IIS-9988642)
%and the Multidisciplinary Research Program of the Department
%of Defense (MURI N00014-00-1-0637). }

% Manual newpage inserted to improve layout of sample file - not
% needed in general before appendices/bibliography.

\newpage

\vskip 0.2in

\end{document}